\documentclass[journal]{IEEEtran}

\ifCLASSINFOpdf
\else
\fi

\usepackage{graphics} 
\usepackage{epsfig} 
\usepackage{mathptmx} 
\usepackage{times} 
\usepackage{amsmath} 
\usepackage{amssymb}  
\usepackage{graphicx}
\usepackage{epstopdf}
\usepackage{hyperref}
\usepackage{makecell}
\usepackage{graphics,graphicx,amssymb,amsmath,verbatim}
\usepackage{mathrsfs}
\usepackage{amsfonts}
\usepackage{epstopdf}
\usepackage{xcolor}
\usepackage{color}
\usepackage{subfigure}
\usepackage{hyperref}
\usepackage{booktabs}
\usepackage{threeparttable}

\usepackage{algorithm}
\usepackage{algorithmic}
\usepackage{multirow} 
\input{tcilatex}
\hypersetup{
colorlinks=true,
linkcolor=black,
citecolor=black
}
\newenvironment{proof of Proposition 1}[1][Proof of Proposition 1]{\textbf{\noindent{\textit{#1.}}} }{\ \rule{0.5em}{0.5em}}

\newenvironment{Remark 1}[1][Remark 1]{{\textit{#1.}} }{\ }
\newenvironment{proof of Theorem 2}[1][Proof of Theorem 2]{\textbf{\noindent{\textit{#1.}}} }{\ \rule{0.5em}{0.5em}}
\newenvironment{proof of Theorem 3}[1][Proof of Theorem 3]{\textbf{\noindent{\textit{#1.}}} }{\ \rule{0.5em}{0.5em}}
\newenvironment{proof of Theorem 4}[1][Proof of
Theorem 4]{\textbf{\noindent{\textit{#1.}}} }{\ \rule{0.5em}{0.5em}}
\newenvironment{proof of Theorem 5}[1][Proof of Theorem 5]{\textbf{\noindent{\textit{#1.}}} }{\ \rule{0.5em}{0.5em}}
\newenvironment{evaluation metrics}[1][Evaluation Metrics]{\textbf{\noindent{\textit{#1.}}} }{}
\title{\Large \bf
Observability-Aware Active Calibration of Multi-Sensor Extrinsics for Ground Robots via Online Trajectory Optimization
}

\author{Jiang Wang, Yaozhong Kang, Linya Fu, Kazuhiro Nakadai,~\IEEEmembership{Fellow,~IEEE,} and He Kong
\thanks{The work of Jiang Wang, Yaozhong Kang, Linya Fu, and He Kong was supported by the National Key R\&D Program of China under Grant No. 2024YFB4710902, the National Natural Science Foundation of China (NSFC) under Grant No. U24A20265, the Science, Technology, and Innovation Commission of Shenzhen Municipality, China, under Grant No. ZDSYS20220330161800001, the Shenzhen Science and Technology Program under Grant No. KQTD20221101093557010, the Guangdong Science and Technology Program under Grant No. 2024B1212010002. Jiang Wang's work was also supported by the JST BOOST program under Grant No. JPMJBS2430. Corresponding author: He Kong.}
\thanks{Jiang Wang and Yaozhong Kang contributed equally to this work. Jiang Wang, Yaozhong Kang, Linya Fu, and He Kong are with the Shenzhen Key Laboratory of Control Theory and Intelligent Systems, and the Guangdong Provincial Key Laboratory of Fully Actuated System Control Theory and Technology, at the Southern University of Science and Technology (SUSTech), Shenzhen 518055, China; Email: wangjiang@ra.sc.eng.isct.ac.jp, kangyz2021@mail.sustech.edu.cn, 12232297@mail.sustech.edu.cn, kongh@sustech.edu.cn. Kazuhiro Nakadai is with the Department of Systems and Control Engineering, Institute of Science Tokyo, Tokyo, Japan; Email: nakadai@ra.sc.eng.isct.ac.jp}
\thanks{Major parts of this work were completed when Jiang Wang was at SUSTech; he is now with the Department of Systems and Control Engineering, Institute of Science Tokyo (formerly Tokyo Tech), Tokyo, Japan.}
}

\begin{document}

\maketitle

\begin{abstract}
Accurate calibration of sensor extrinsic parameters for ground robotic systems (i.e., relative poses) is crucial for ensuring spatial alignment and achieving high-performance perception. However, existing calibration methods typically require complex and often human-operated processes to collect data. Moreover, most frameworks neglect acoustic sensors, thereby limiting the associated systems' auditory perception capabilities. To alleviate these issues, we propose an observability-aware active calibration method for ground robots with multimodal sensors, including a microphone array, a LiDAR (exteroceptive sensors), and wheel encoders (proprioceptive sensors). Unlike traditional approaches, our method enables active trajectory optimization for online data collection and calibration, contributing to the development of more intelligent robotic systems. Specifically, we leverage the Fisher information matrix (FIM) to quantify parameter observability and adopt its minimum eigenvalue as an optimization metric for trajectory generation via B-spline curves. Through planning and replanning of robot trajectory online, the method enhances the observability of multi-sensor extrinsic parameters. 
The effectiveness and advantages of our method have been demonstrated through numerical simulations and real-world experiments. For the benefit of the community, we have also open-sourced our code and data at https://github.com/AISLAB-sustech/Multisensor-Calibration.
\end{abstract}


\section{Introduction}
Precise calibration of extrinsic parameters of robotic systems, namely, the relative positions and orientations between sensors, is essential for achieving accurate spatial alignments and effective multimodal sensor fusion \cite{Heckman19}-\cite{ex2}. 
Moreover, sensor parameters inevitably drift over time due to factors such as environmental noises and system vibrations. If a robot continues its operations without proper calibration (or re-calibration) under these scenarios, it risks severe performance degradation or even system/mission failures. Therefore, ensuring accurate and timely calibration of sensor parameters has remained an important topic in the robotics and automation community (see \cite{Maye16}-\cite{Rehder16} and the references therein for some recent progress).

In practice, as a complement to vision and other types of sensors, microphone arrays are widely used for robot audition tasks such as sound source localization \cite{Fu2024}, detection \cite{Cao24}, recognition \cite{Chen24}, and tracking \cite{AnTRO}. Acoustic sensors significantly enhance the robot's ability to perceive and interact with its environment \cite{Grondin2022}-\cite{Nakadai2015}. 
However, acoustic sensors are neglected in most existing works on multimodal sensor calibration \cite{Maye16}-\cite{Rehder16}. This omission hampers the robot's ability to integrate useful auditory data with other modalities \cite{Yang20}.
\begin{figure}[t]
\centering 
{\includegraphics[width=0.85\columnwidth]{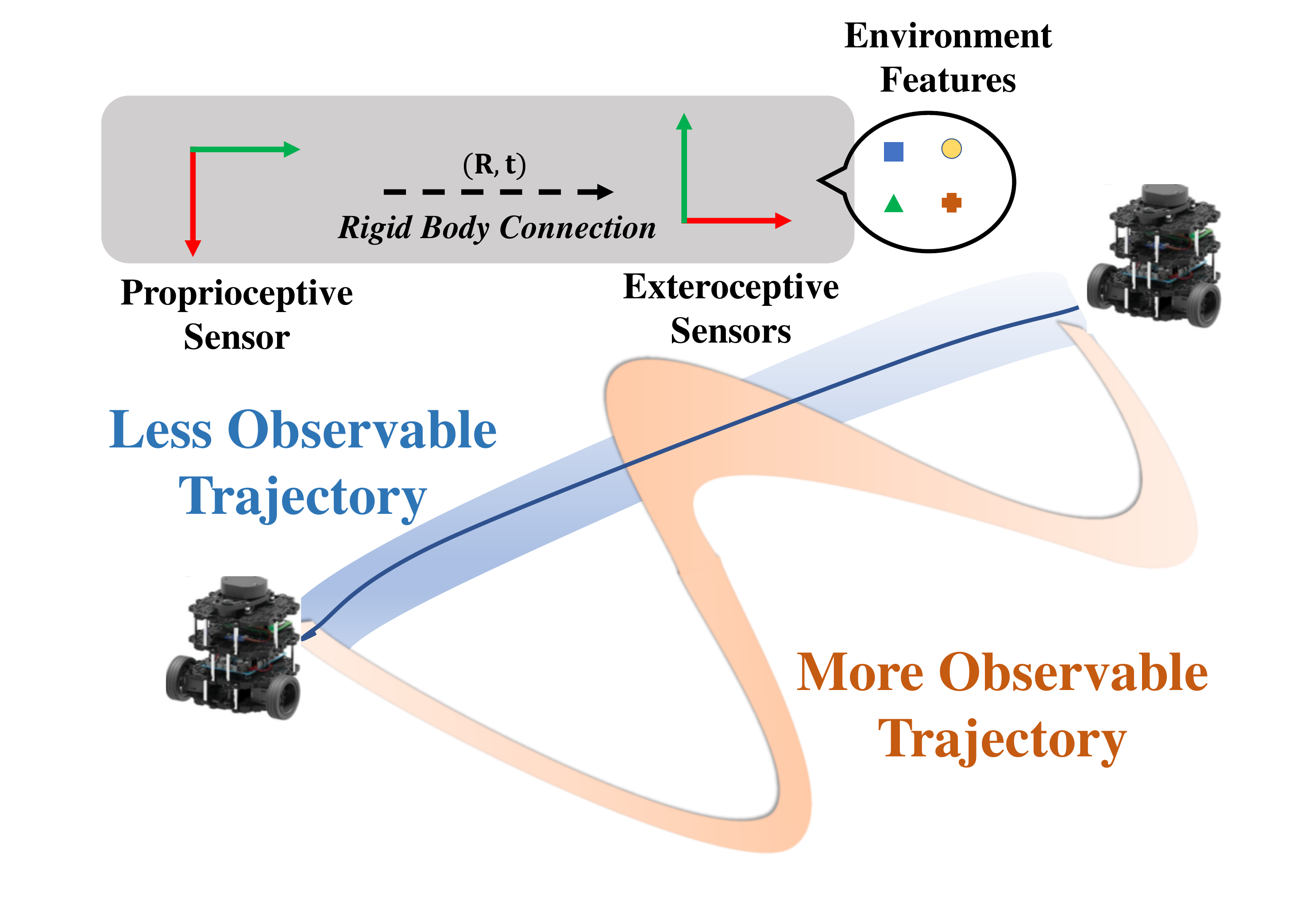}
}
\caption{Online trajectory planning for mobile robots to actively calibrate sensors extrinsic parameters.}
\label{planning} 
\end{figure}



Moreover, most existing calibration techniques rely on manually collected data, which is then processed by offline calibration algorithms to estimate the sensor parameters \cite{Heckman19,ex3,ex4}. Although some recent studies have explored the concept of observability (i.e., whether the available measurements contain enough information for identification of unknown parameters) for improved accuracy \cite{Mirzaei08}-\cite{Teller12}, they typically require experienced operators to guide the robot along trajectories with sufficient motion excitation to make the system parameters observable. This requirement is difficult to achieve for robotic platforms operating in the field, or for users who lack expertise in calibration procedures.

To overcome the above limitations, active calibration for multi-sensor extrinsic parameters is an important alternative. In robotics, active sensing or active information acquisition is defined as the problem of managing the robot’s trajectories to maximize the information
gathered regarding a target system/process \cite{Bajcsy18}. 
Similarly, active calibration refers to automated data collection during the calibration process according to specific optimization metrics \cite{Schneider19}.
As illustrated in Fig. 1, improving data collection largely depends on planning the robot's future trajectories that can maximize the information provided to the model which, in turn, raises the question of how to evaluate the informativeness of these measurements.

The Fisher information matrix (FIM) approach is a widely-used framework for system identification and parameter estimation in the robotics and automation community (see, e.g., \cite{KongAuto2023}-\cite[chap. 2]{Barfoot} and the references therein). From the information-theoretic perspective, the FIM quantifies the amount of information contained in a set of observations about the unknown parameters \cite{Dissanayake2008}. In this paper, by leveraging the FIM approach, we propose an observability-aware active calibration method for the extrinsic parameters of multimodal sensors (i.e., a microphone array, along with LiDAR and wheel odometry) for ground robotic systems. Specifically, the main contributions of this work are as follows:
\begin{enumerate}
    \item We propose an active calibration framework that enhances parameter observability through online trajectory planning and replanning. By leveraging FIM and maximizing its minimum eigenvalue, the framework optimizes the robot's trajectory to improve data collection for parameter estimation. Compared to widely-used non-optimized trajectories, this approach significantly reduces uncertainty and enhances calibration accuracy. 
    \item To address synchronization challenges among multimodal sensors, we leverage B-spline curves to represent continuous trajectories. This method allows seamless interpolation of sensor data and avoids complexities caused by varying sampling rates. By optimizing the control points of the B-spline curves, we efficiently generate calibration trajectories. An extended Kalman filter (EKF) is used to perform real-time extrinsic calibration across all sensor modalities.
    \item Extensive simulation and real-world studies have been carried out to validate and illustrate the effectiveness and merits of the proposed method. Our results demonstrate significant improvements in parameter estimation accuracy over standard trajectories, such as figure-8 and circular paths. Additionally, we have open-sourced the code to benefit the broader robotics community at https://github.com/AISLAB-sustech/Multisensor-Calibration.
\end{enumerate}

\textbf{Notation}: Denote $x$, $\mathbf{x}$, and $\mathbf{X}$ as
scalars, vectors, and matrices, respectively. $\mathbf{X}^{\top}$ represents the transpose of matrix $\mathbf{X}$. $\mathbb{R}^{n}$ denotes the $n$-dimensional
Euclidean space. $[a_{1};\cdots;a_{n}]$ denotes $[a_{1}^{\top},\cdots,a_{n}^{\top}]^{\top}$,
where $a_{1},\cdots,a_{n}$ are scalars/vectors/matrices with proper
dimensions. $diag_{n}(a)$ denotes a  diagonal matrix with $a$ as element diagonal entries for $n$ times; $\left\Vert\mathbf{a}\right\Vert_2$ denotes the square norm of vector $\mathbf{a}$; $diag(\mathbf{A},\mathbf{B})$
denotes a block diagonal matrix with $\mathbf{A}$ and $\mathbf{B}$ as
its block diagonal entries;  $\mathbf{0}$ stands for a matrix of appropriate dimensions with its all entries as 0. Vectors/matrices, with dimensions not explicitly stated, are assumed to be algebraically compatible.

\section{Related Work}
Based on the data acquisition methods, existing works in extrinsic calibration of multi-sensors 
can be broadly categorized into two types: passive calibration and active calibration. Different from passive calibration where datasets are usually collected via manual operation, active calibration allows the multisensor platform to autonomously gather necessary data based on optimization \cite{Schneider19}. As mentioned earlier, most existing works have overlooked microphone arrays. Hence, for the sake of completeness, we next first review advancements in microphone array calibration, followed by a discussion on passive and active calibration methods for multimodal sensors.

\subsection{Microphone Array Calibration}
In the past couple of decades, the calibration of single microphone arrays has been extensively studied. A calibration algorithm based on time difference of arrival (TDOA) between each pair of microphones was proposed in \cite{Perrodin 2012}. In \cite{Crocco 2011}, a bilinear calibration method based on time of flight (TOF) between sensors and sound sources was introduced, allowing the estimation of sensors and sound sources positions under known emission times. Similarly, \cite{Kuang 2013} proposed a joint calibration framework for determining microphone positions and source locations using time of arrival (TOA) measurements, assuming prior knowledge of distances between sources and microphones.

For the calibration of multiple microphone arrays, \cite{Plinge2017} utilized direction of arrival (DOA) to determine sound source locations, while TDOA measurements were processed through evolutionary algorithms to estimate microphone array poses. Additionally, \cite{Yin 2021} presented a distributed damped Newton optimization framework for calibrating multiple microphone arrays based on DOA and inter-array TDOA measurements. In \cite{Wang 2021}, a 3D calibration approach employing artificial bee colony algorithms was proposed under the assumption that the source positions at certain times are partially known.

However, most existing methods assume synchronized microphone channels, which limits their practical applicability. To address this, in \cite{Kong2021}, we have proposed a general framework based on batch simultaneous localization and mapping (SLAM) for joint source localization and calibration of a single microphone array with asynchronous effects (i.e., clock difference and initial time offset). 

To further improve calibration accuracy, our recent works, as presented in \cite{Zhang24} and \cite{Li24}, proposed to combine different types of TDOA measurements and use mutual information to identify the most informative data segments of measurements, respectively. Furthermore, the SLAM-based framework in \cite{Kong2021} has been extended to the case with multiple microphone arrays in \cite{Jiang24}.


Despite these advancements, existing techniques predominantly treat microphone arrays as an isolated sensing modality, neglecting their integration with other sensors. This limitation hinders the effective fusion of auditory data with other modalities in robotic systems, restricting their ability to perform multimodal perception tasks effectively.

\subsection{Passive Multimodal Sensor Calibration}
Passive calibration of multimodal sensors has been extensively explored in robotic research. In \cite{Mirzaei08}, a Kalman filter-based approach was introduced to estimate the relative pose between a camera and an IMU. During the calibration, the sensors were randomly moved in front of a chessboard to sufficiently excite the system. This allowed for offline parameter estimation by coupling the camera-captured images of the board with the motion recorded by the IMU. \cite{Kelly10} introduced the concept of self-calibration, defined as the use of measurements solely from the sensors themselves to improve estimates of system parameters. Although they still utilized a chessboard as a visual feature during calibration, no prior knowledge of the board attributes was provided.

\cite{Rehder16} formulated the multimodal sensor calibration problem as a batch optimization task. 
and validated this approach on a setup comprising a stereo camera and a laser rangefinder. In \cite{Teller12}, observability analysis based on the FIM revealed that the extrinsic parameters of multimodal sensors were unobservable in certain specific scenarios. \cite{Censi13} proposed a method to jointly calibrate a laser rangefinder and odometry by formulating the problem as a maximum likelihood estimation task, under the assumption that measurements were collected along adequately exciting trajectories.

Under similar assumptions, \cite{Maye16} leveraged the algebraic link between the Gauss-Newton algorithm, the FIM, and nonlinear observability analysis to automatically detect numerically unobservable directions in the parameter space, preventing updates along these directions. 
Later, \cite{Schneider19} reevaluated the information metrics to select high-information subsequences from long trajectories for calibration and applied the approach to the camera and inertial platform. \cite{LIU 2022} adapted the method from \cite{Maye16} to calibrate the LiDAR and inertial platform, ensuring that observable parameters could still be calibrated even in degraded scenarios lacking sufficient excitation. Very recently, \cite{Woosik2024} integrated various sensor types, including IMU, camera, wheel encoder, GPS, and LiDAR, into a simultaneous online calibration framework based on the multi-state constraint Kalman filter (MSCKF). 


There exist a few works that consider calibration of systems with both vision and acoustic sensors. For example, \cite{Legg13} proposed to align the positions of sound sources in the microphone array frame with their locations on the chessboard for acoustic imaging applications. Similarly, for underwater systems with sonar and cameras, \cite{Yang20} has proposed framework based on the well-known Perspective-n-Point algorithm.
Later efforts by \cite{Wang20} improved the design of calibration board features to enhance the precision of image data acquisition in turbid underwater environments. However, the above-mentioned passive calibration methods lack the ability to influence robot motion during the data collection process. When there is insufficient motion across all degrees of freedom, these methods may result in inaccurate extrinsic calibration.

\subsection{Active Multimodal Sensor Calibration}
In contrast to passive calibration methods, active calibration involves planning and executing trajectories to actively collect data required for sensor calibration \cite{Schneider19}. Active calibration has received increasing attention in recent years. For instance, \cite{Murali16} proposed a greedy planner that selects actions from a predefined set of simple movements to minimize uncertainty in sensor position parameters during each robot motion. However, their approach did not account for sensor orientation estimation. \cite{Preiss18} introduced the expanded empirical local observability Gramian to approximate the local observability Gramian, reducing computational complexity; this method was used to generate trajectories that avoid unobservable subspaces, and its effectiveness was demonstrated on systems comprising GPS, camera, and IMU. 

Building on the work \cite{Maye16}, \cite{Heckman19} introduced a reinforcement learning-based framework to determine motion sequences that maximize inference of the desired parameters. This approach was validated on a platform with camera and IMU. \cite{Shida23} employed the FIM to assess system observability; when extrinsic parameters were partially unobservable, they actively searched for the next optimal motion using entropy-based optimization to restore observability.

In this study, we propose a novel observability-aware multimodal sensor calibration method that jointly calibrates microphone arrays, LiDAR, and odometry for ground robotic systems. Our motivation in this work is in close spirit to the above-mentioned works, e.g., \cite{Heckman19} and \cite{Shida23}, in that we quantify system observability through the FIM. Although the methods in \cite{Heckman19} and \cite{Shida23} can be applied for a more diverse range of platforms such as air or underwater robots, \cite{Heckman19} only selects predefined motions from a library of designed actions, while \cite{Shida23} determines the next best motion without considering the full trajectory. 

In contrast, our method combines the FIM with B-spline curves to provide a more flexible and comprehensive active calibration framework. This approach generates observability-aware trajectories online to collect data and calibrate sensor parameters in real-time. To the best of our knowledge, this is the first work to jointly calibrate the microphone array with proprioceptive sensors in multimodal systems, a gap that we aim to fill in this work. 

\section{Problem Formulation}
\subsection{Problem Statement}
We consider a robot system in a two-dimensional environment equipped with multiple sensors, including exteroceptive sensors (microphone array and LiDAR) and proprioceptive sensor (wheel odometry). As shown in Fig. \ref{planning}, each sensor is assumed to have a local frame attached to it, and the global frame is chosen to coincide with the wheel odometer frame at time instance $k=0$, denoted as $\left\{ \mathbf{G}\right\}$. During the calibration, it is assumed that all sensors remain rigidly attached to the robot and the wheel odometer frame is treated as the robot frame. 

At any time instance $k$, where $k=1,\ldots,K$, $^{G}\mathbf{R}_{k}$ is the rotation matrix of the global frame $\left\{ \mathbf{G}\right\}$ to the wheel odometer frame $\left\{ \mathbf{W}_{k}\right\}$ with the orientation $^{G}\theta_{k}$, and $^{G}\mathbf{t}_{k}$ is the wheel odometer position in the global frame. Then, the pose of the LiDAR frame $\left\{ \mathbf{L}_{k}\right\} $ in the global frame can be represented by the rotation matrix $^{W}\mathbf{R}_{L}$ (with a rotation angle of $^{W}\theta_{L}$ relative to the wheel odometer) and the translation vector $^{W}\mathbf{t}_{L}$ with respect to (w.r.t.) the wheel odometer pose \cite[Ch. 3]{Lynch2017}. Similarly, the pose of the microphone array frame $\left\{ \mathbf{M}_{k}\right\} $ in the global reference frame can be represented by the rotation matrix $^{W}\mathbf{R}_{M}$ (with a rotation angle of $^{W}\theta_{M}$ relative to the wheel odometer) and the translation vector $^{W}\mathbf{t}_{M}$. Denote the microphone array and LiDAR extrinsic parameters $\mathbf{\psi}_M$ and $\mathbf{\psi}_L$ as
\begin{equation}
\mathbf{\psi}_M=\left[^{W}\mathbf{t}_{M};^{W}\theta_{M}\right],\text{ }\mathbf{\psi}_L=\left[^{W}\mathbf{t}_{L};^{W}\theta_{L}\right],
\end{equation}
respectively. Thus, all sensors extrinsic parameters $\mathbf{\psi}$ to be identified are:
\begin{equation}
\mathbf{\psi}=\left[\mathbf{\psi}_M;\mathbf{\psi}_L\right].
\end{equation}
By using the standard notation of SLAM \cite{Huang2016}, we define the robot's system model as:
\begin{equation}
\begin{aligned}
 \mathbf{x}_k= & \mathbf{f}\left(^{G}\mathbf{t}_{k-1},^{G}\theta_{k-1},\mathbf{u}_{k},\mathbf{\psi}\right)\\
\mathbf{z}_k= & \mathbf{g}\left(^{G}\mathbf{t}_{k},^{G}\theta_{k},\mathbf{\psi}\right)+\mathbf{w}_{k}
\end{aligned},\label{eq:system}
\end{equation}
where $\mathbf{x}_{k}$, $\mathbf{u}_{k}$, and $\mathbf{z}_{k}$ stand for the system state, the control input, and the sensor measurements, at the $k$-th sampling instant, respectively; $\mathbf{w}_{k}\sim\left(\mathbf{0},\mathbf{N}_{k}\right)$ is a Gaussian-distributed observation noise vector, with known covariance $\mathbf{N}_{k}$;  $\mathbf{f}(\cdot)$ and $\mathbf{g}(\cdot)$ represent the robot state transition function and the measurement function, respectively.

Given the robot system model described above, our primary objective is (1) to derive a maximum a posteriori (MAP) estimate $\hat{\psi}$ for the sensor parameters
\begin{equation}
\hat{\psi}=\underset{\psi}{\arg\max}p\left(\psi\mid\mathbf{z}_{1:K},\mathbf{u}_{1:K}\right)
\end{equation}
from the robot's control inputs and sensor measurements,
and (2) improve the sensor parameter estimation accuracy by actively planning the robot trajectory online.


\subsection{Sensor Measurements and Models}
\subsubsection{\textbf{Wheel Odometry}}
Wheeled robots are typically equipped with an encoder on each wheel, providing local angular velocity readings. These readings can be pre-integrated to obtain a series of wheel odometer measurements over time, yielding the relative translation $\Delta\mathbf{t}_{k-1}^{k}$ and the relative orientation $\Delta \theta_{k-1}^{k}$ of the odometer frame, represented as
\begin{equation}
\left[\begin{array}{c}\Delta\mathbf{t}_{k-1}^{k}\\
\Delta \theta_{k-1}^{k}
\end{array}\right]=
\left[\begin{array}{c}
\intop_{k-1}^{k}\dfrac{w_{l}^t+w_{r}^t}{2}r\cos^{G}\theta_{t}dt\\
\intop_{k-1}^{k}\dfrac{w_{l}^t+w_{r}^t}{2}r\sin^{G}\theta_{t}dt\\
\intop_{k-1}^{k}\dfrac{-w_{l}^t+w_{r}^t}{d_{w}}rdt
\end{array}\right],
\end{equation}
where $w_{l}^t$ and $w_{r}^t$ are the angular velocities of the left and right wheels, respectively, $r$ is the wheel radii, and $d_{w}$ is the length of the wheelbase. Therefore, the pose of the wheel odometer $^{G}\mathbf{t}_{k}$ and $^{G}\theta_{k}$ at time instance $k$ can be obtained through the accumulation of $\Delta\mathbf{t}_{k-1}^{k}$ and $\Delta \theta_{k-1}^{k}$.
\subsubsection{\textbf{Microphone Array}}
When the geometry of an array with $M$ microphones is known, the direction of arrival (DOA) of a sound source relative to the microphone array frame can be accurately estimated by processing the signal $\Phi_k$ captured by the microphones. In this paper, a grid search method based on the steered response power with phase transform (SRP-PHAT) algorithm is used to estimate the DOA. It is widely used in robotic sound source localization, and the phase transform enhances localization performance in reverberant environments \cite{Fu2024,AnTRO}.

For each potential sound source direction of arrival $\alpha_k$, the cumulative signal power $Pow(\alpha_k,\Phi_k)$ of the different microphone pairs is calculated as 
\begin{equation}
Pow\left(\alpha_k,\Phi_k\right)=\sum_{m=1}^{M-1}\sum_{n=m+1}^{M}\int_{-\infty}^{\infty}\dfrac{\varphi_{m}(f)\varphi^{*}_{n}(f)}{\left\Vert\varphi_{m}(f)\varphi^{*}_{n}(f)\right\Vert_2}e^{j2\pi f\tau_{m,n}(\alpha_k)}df
\end{equation}
where $\varphi_(f)$ is the discrete Fourier transform of the microphone signals at the frequency $f$, and $\varphi_{m}(f)\varphi^{*}_{n}(f)$ represents the cross power spectrum ($*$ denotes the complex conjugate) \cite{jasa22}; $\tau_{m,n}(\alpha_k)$ is the theoretical TDOA for the microphone pair $(m,n)$ w.r.t. the potential sound source direction $\alpha_k$. The DOA estimation $\hat{\alpha}_k$ is determined as the angle corresponding to the maximum power with phase transformation, expressed as
\begin{equation} 
\begin{array}{c}
\hat{\alpha}_k=\underset{\alpha_k}{\arg\max}Pow\left(\alpha_k,\Phi_k\right).\end{array}
\end{equation}
 The vector $\mathbf{d}_{k}$, corresponding to the DOA $\hat{\alpha}_k$ obtained from signal processing at time instance $k$, can be modeled geometrically in space as
\begin{equation}
\mathbf{d}_{k}=\left(^{G}\mathbf{R}_{k}\cdot^{W}\mathbf{R}_{M}\right)^{\top}\frac{\mathbf{s}-\left(^{G}\mathbf{t}_{k}+^{G}\mathbf{R}_{k} \cdot ^{W}\mathbf{t}_{M}\right)}{\left\Vert \mathbf{s}-\left(^{G}\mathbf{t}_k+^{G}\mathbf{R}_{k} \cdot ^{W}\mathbf{t}_{M}\right)\right\Vert_2 }
\label{expression_DOA-1}
\end{equation}
where $\mathbf{s}$ is the known sound source position in the global frame. This measurement model establishes a correlation of the relative pose between the microphone array and the wheel odometry.

\subsubsection{\textbf{LiDAR}}
Mechanical LiDAR, which is widely used, emits laser beams as the laser head rotates around a central mechanical axis. It receives reflected signals to determine the distance and bearing of the scanned points. Since LiDAR generates a large number of data points, real-time tracking of all points is challenging \cite{fastlio}. A more practical approach is to accumulate points over a certain period and process them collectively. In this paper, we use the line features from LiDAR scans to align point cloud data between two frames, thereby solving for the translation and orientation of the current frame relative to the previous one.

To be more specific, one needs to first determine the correspondence between two data frames $P_{k-1}$ and $P_{k}$ (i.e., data association) consisting of $N$ scan points. For each point $\mathbf{p}_{k-1}^i$ in the data frame $P_{k-1}$, nearest neighbor search is used to find the two points $\mathbf{p}_{k}^a$ and $\mathbf{p}_{k}^b$ in the current frame $P_{k}$, forming a line segment ${\mathbf{p}_{k}^{a}\mathbf{p}_{k}^{b}}$ \cite{Pomerleau F}. Then, to estimate the LiDAR pose change $\Delta \mathbf{\theta}_{L,k}$ and $\Delta \mathbf{t}_{L,k}$ between two data frames, the distance between each point in the current frame and the line defined by its two nearest neighbors is minimized, resulting in the following nonlinear optimization problem:
\begin{equation}
\underset{\Delta \mathbf{\theta}_{L,k},\Delta \mathbf{t}_{L,k}}{\min}\sum_{i=1}^{N}D\left(\mathbf{p}_{k}^{i}{\prime} ,\mathbf{p}_{k}^{a}\mathbf{p}_{k}^{b}\right)^{2}
\end{equation}
where the distance function ${D}(\mathbf{p}_{k}^{i}{\prime},\mathbf{p}_{k}^{a}\mathbf{p}_{k}^{b})$ is given by
\begin{equation}
{D}(\mathbf{p}_{k}^{i}{\prime},\mathbf{p}_{k}^{a}\mathbf{p}_{k}^{b})=\frac{\left\Vert\left(\mathbf{p}_{k}^{b}-\mathbf{p}_{k}^{a}\right)\times\left(\mathbf{p}_{k}^a-\mathbf{p}_{k}^i{\prime}\right)\right\Vert_2}{\left\Vert\mathbf{p}_{k}^b-\mathbf{p}_{k}^a\right\Vert_2},
\end{equation}
and 
\begin{equation}
\mathbf{p}_{k}^{i}{\prime} = \mathbf{R}\left(\Delta \mathbf{\theta}_{L,k}\right)\mathbf{p}_{k-1}^i+\Delta \mathbf{t}_{L,k},
\end{equation}
where $\mathbf{R}\left(\Delta \mathbf{\theta}_{L,k}\right)$ is the rotation matrix of the LiDAR frame from $\left\{ \mathbf{L}_{k-1}\right\}$ to $\left\{ \mathbf{L}_{k}\right\}$ with $\Delta \mathbf{\theta}_{L,k}$. 
As shown in the existing literature \cite{Censi08}, the problem above has a closed-form solution that can be used as the initial value for further improvements via more advanced optimization. The pose change between two LiDAR frames can be modeled in space as
\begin{equation}
\begin{array}{c}
\Delta\mathbf{t}_{L,k}=\left(^{G}\mathbf{R}_{k-1}\cdot^{W}\mathbf{R}_{\mathit{L}}\right)^{\top}
\left(\Delta^{G}\mathbf{R}_{W,k}\cdot^{W}\mathbf{t}_{L}+\Delta\mathbf{t}_{k-1}^{k}
\right)\\
\Delta \theta_{L,k}=^{G}\theta_{k}-^{G}\theta_{k-1}
\end{array},
\label{expression_lidar}
\end{equation}
where 
\begin{equation}
\begin{array}{c}
\Delta^{G}\mathbf{R}_{W,k} = ^{G}\mathbf{R}_k-^{G}\mathbf{R}_{k-1},\text{ } 
\Delta\mathbf{t}_{k-1}^{k} = ^{G}\mathbf{t}_{k}-^{G}\mathbf{t}_{k-1}.
\end{array}
\label{lidarglobal}
\end{equation}
The established model above associates the pose changes of the wheel odometry and LiDAR between two sampling instances.

\begin{figure}[t]
\centering {\includegraphics[width=0.95\columnwidth]{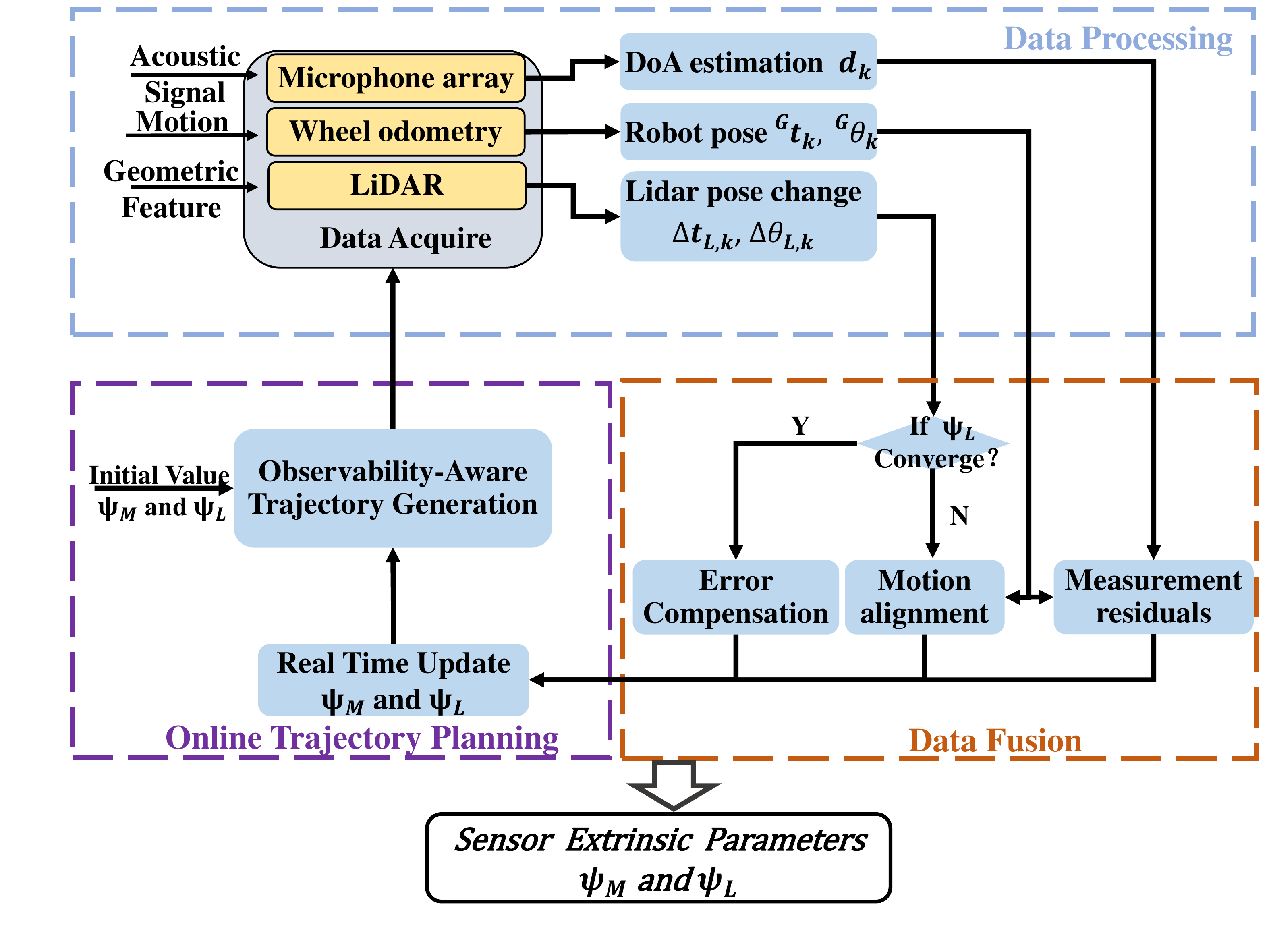}}
\caption{Pipeline of the proposed calibration method.}
\label{Overview} 
\end{figure}

\section{\label{pre-1}Observability-Aware Active Calibration}

\subsection{System Overview}
Fig. \ref{Overview} illustrates the workflow of the proposed active extrinsic calibration method for multi-sensor systems, which comprises the following steps:

\subsubsection{\textbf{Initial Trajectory Generation}} Based on the initial sensor parameters (e.g., manual measurements or empirical estimations) and the observation models of each sensor, the FIM is utilized to quantify and optimize the observability of sensor parameters, resulting in an initial global calibration trajectory composed of multi-segmented B-spline curves.

\subsubsection{\textbf{Data Processing and Real-Time Extrinsic Updating}} The robot follows the generated trajectory under a designed trajectory tracking controller (e.g., PID), while sensors collect data and update the sensor extrinsic parameters in real-time. The microphone array estimates the DOA of the sound source by identifying the maximum signal power in each audio frame. Measurement residuals calculated as the difference between the predicted and theoretical values are input into an extended Kalman filter to refine the extrinsic parameters. For LiDAR and the wheel odometer, the relative pose changes are estimated by aligning line features from adjacent point cloud frames and integrating travel distance, respectively. These measurements are used to optimize LiDAR extrinsic parameters via trajectory alignment. The EKF combines sensor predictive models with measurements to perform prediction and update operations, refining the extrinsic parameters and their covariance matrix to reduce estimation uncertainty and enhance calibration accuracy.

\subsubsection{\textbf{Error Compensation}} 
Despite its short-term accuracy, wheel odometry often suffers from rapid drift in practical environments due to wheel slippage and uneven surfaces, rendering its pose estimates unreliable after only a few meters of motion \cite{Cadena2016}. In contrast, LiDAR can produce more accurate pose estimates over longer durations by leveraging rich geometric features in the environment \cite{Censi08}. Moreover, as demonstrated in Section VI, the LiDAR parameters typically converge faster and more accurately than those of the microphone array. Inspired by these observations, we leverage the high-precision pose estimates from LiDAR to further improve the calibration performance of the microphone array. Specifically, once the update variance of the LiDAR parameters falls below a predefined threshold, the system transforms the LiDAR pose estimate into the robot frame and substitutes it for wheel odometry. This replacement provides more reliable robot pose estimation and thus facilitates more accurate array calibration.



\subsubsection{\textbf{Trajectory Replanning Online}} When the robot approaches the end of the current trajectory segment, Step (1) is repeated with the updated extrinsic parameters to optimize the subsequent trajectory (i.e., the following B-spline segments) until the system converges (when the update step size is smaller than the set threshold) or the robot reaches the end of the last segment.

In the next few subsections, we will discuss the core mechanisms related to trajectory generation and updating (Step 4 as described above) for the proposed system. A summary of this algorithm is provided in Algorithm 1.

\subsection{Continuous Trajectory Representation}
We use a parameterization approach, namely, clamped B-spline curves for continuous trajectory representation. We choose the above framework for the following reasons: (i) it allows for evaluating the trajectory at any point, ensuring smoothness and continuity, and enabling the derivation of theoretical values of the sensor measurements at any sampling instant; (ii) it affords a closed-form derivative, and allows for the imposition of additional motion constraints, and is particularly advantageous for fusing asynchronous multi-sensor measurements (i.e., data streams with different sampling frequencies); (iii) it enables adjusting specific segments of the trajectory by modifying a limited number of control points, leading to efficient trajectory updates and optimization. 

To be more specific, a $d-$degree clamped B-spline curve consists of $n-d$ segments polynomial spline curves, where $n$ is the number of control points and $n > d$. For $j=1,2,\ldots, n-d$, the polynomial piece trajectory
$\mathbf{T}_{j}$ 
is generated by fitting the control point set $\mathbf{C}_j=\left[\mathbf{c}_{j};\mathbf{c}_{j+1};\ldots;\mathbf{c}_{j+d}\right]$. 
Then, any point on the continuous trajectory $\mathbf{T}_{j}$ can be represented as
\begin{equation}
\mathbf{T}_{j}(q)=\mathbf{q}^{\top}\mathbf{M}_{j}^{(d+1)}\mathbf{C}_j,
\end{equation}
where $\mathbf{q}=\left[1;q;\ldots;q^{d}\right]$ and $q \in\left(0,1\right]$. A smaller $q$ indicates proximity to the beginning of the segment, while a larger $q$ signifies closeness to the segment end. $\mathbf{M}_{j}^{(d+1)}\in \mathbb{R}^{(d+1) \times (d+1)}$ is the corresponding cumulative basis function matrix \cite{splinematrix1}, which depends on the degree of the spline curve and the index of the corresponding trajectory segment. In this paper, a cubic B-spline $(d=3)$ is used to represent trajectories, ensuring both computational efficiency in trajectory generation and continuity in the mobile robot acceleration during motion.

\subsection{Observability via the FIM Approach}
In robotic applications, the observability of a system is defined as the ability to infer the unknown states/parameters of the system from a sequence of measurements and is often assessed through the FIM approach \cite{Preiss18, Huang2016}. A system is considered observable if the FIM is non-singular, i.e., its rank equals the number of unknown states \cite{Teller12,Kong2021}. 

To be more specific, the FIM of the system in (\ref{eq:system}) for the trajectory segment $\mathbf{T}_{j}$ can be constructed as 
\begin{equation}
\mathbf{I}_{FIM\_j}=\mathbf{J}_{j}^{\top}\left(\psi\right){\Sigma}_j^{-1}\mathbf{J}_{j}\left(\psi\right),
\end{equation}
where $\mathbf{J}_{j}\left(\psi\right)$ is the cumulative Jacobian matrix of the observation model $\mathbf{g}$ in (\ref{eq:system}) w.r.t. the unknown parameters $\psi$ at the sampling positions within the segment $\mathbf{T}_{j}$, and can be written as
\begin{equation}
\begin{array}{c}
\mathbf{J}_{j}\left(\psi\right)= \left[\begin{array}{cccc}
\frac{\partial \mathbf{g}_l}{\partial\psi};\frac{\partial \mathbf{g}_{l+1}}{\partial\psi}; \ldots; \frac{\partial \mathbf{g}_r}{\partial\psi}\end{array}\right]\\ 
=\begin{bmatrix}
\begin{smallmatrix}
\frac{\partial\mathbf{d}_{l}}{\partial\psi_{M}}^{\top} & \mathbf{0} & \mathbf{0} & \frac{\partial\mathbf{d}_{l+1}}{\partial\psi_{M}}^{\top} & \mathbf{0} & \mathbf{0} & \ldots & \frac{\partial\mathbf{d}_{r}}{\partial\psi_{M}}^{\top} & \mathbf{0} & \mathbf{0}\\
\mathbf{0} & \frac{\partial\Delta\mathbf{t}_{L,l}}{\partial\psi_{L}}^{\top} & \mathbf{0} & \mathbf{0} & \frac{\partial\Delta\mathbf{t}_{L,l+1}}{\partial\psi_{L}}^{\top} & \mathbf{0} & \ldots & \mathbf{0} & \frac{\partial\Delta\mathbf{t}_{L,r}}{\partial\psi_{L}}^{\top} & \mathbf{0}
\end{smallmatrix}
\end{bmatrix}^{\top}.
\end{array}
\end{equation}
The detailed expressions of the Jacobian matrix can be found in  Appendix, and $\Sigma_j=diag\left(\mathbf{N}_{l},\mathbf{N}_{l+1},\ldots,\mathbf{N}_{r}\right) $ represents the covariance matrix, and $l, r \in \left[1,K\right]$.

The inverse of the FIM is known as the Cramér-Rao Lower Bound (CRLB), which represents the theoretical lower bound for the parameter estimation error. The minimum eigenvalue of the FIM is closely related to the observability of each parameter \cite[Ch. 2]{Bar-Shalom2004}, i.e., larger minimum eigenvalues correspond to smaller CRLBs, indicating lower variance bounds and uncertainty in parameter estimation, and higher observability of the parameters. Conversely, smaller minimum eigenvalues imply higher variance bounds and lower parameter observability. When the minimum eigenvalue approaches zero, the variance of the parameter estimates approaches infinity, indicating that the parameter is unobservable. Therefore, the minimum eigenvalue of the FIM naturally serves as an indicator for evaluating the observability of parameter estimation.

\subsection{The Optimization Objective}
Trajectories without sufficient excitation to the sensor model usually lead to low observability of the sensor parameters. This is typically reflected by a singular or ill-conditioned FIM, making it impossible to accurately estimate the unknown system parameters \cite{Teller12}. To address this issue, our proposed active calibration framework enables the robot to autonomously generate observability-enhancing trajectories. The observability trajectory generation problem can be formulated as
\begin{equation}\label{opt_Q}
\begin{aligned}\underset{\mathbf{C}}{\mathrm{maximize}\text{ }} & \sigma_{min}\left(\mathbf{I}_{FIM}\right)\\
\text{subject to } & \text{STATES}\left(\mathbf{C}\right) \in \Omega
\end{aligned}
\end{equation}
where $\sigma_{min}\left(\mathbf{I}_{FIM}\right)$ represents the minimum eigenvalue of the FIM for all trajectory segments, which is calculated as
\begin{equation}
\mathbf{I}_{FIM}=\mathbf{J}_{1:n-d}^{\top}\left(\psi\right){\Sigma}_{1:n-d}^{-1}\mathbf{J}_{1:n-d}\left(\psi\right)=\sum_{j=1}^{n-d}\mathbf{I}_{FIM\_j};
\end{equation}
$\mathbf{C}$ denotes the finite set of control points of the B-spline curve, as introduced in Section IV-B, which is derived from the optimization framework and is constrained within the set of all feasible states $\Omega$.

State constraints ensure the feasibility of the planned path. As most robots face non-omnidirectional motion constraints, such as differential or Ackermann steering, additional kinematic constraints $\Omega$ must be incorporated to ensure accurate trajectory tracking by the mobile robot. Specifically, we imposed steering constraints on the planned trajectory, ensuring that the robot steering angle $\beta$ between two consecutive sampling points satisfies $0\leq\beta\leq\beta_{max}$. Additionally, we constrained the robot position within the world frame to remain inside a predefined boundary box, such that $\eta_{min}\leq\mathbf{T}_{j}(q)\leq\eta_{max}$. The optimization problem in (\ref{opt_Q}) can be efficiently solved using the constrained optimization by linear approximations (COBYLA) method as implemented in the Python Optimization Toolbox. 


Note that trajectory planning is an iterative optimization process rather than a one-time computation. The process begins with global path planning based on initial sensor parameters and the measurement models. As the robot follows the path, it incrementally refines the initial parameter estimates using the calibration method described in Section IV-D. Whenever the robot reaches the end of a spline segment, the remaining segments are re-optimized. This iteration continues until the increment step to the sensor parameters falls below a predefined threshold or the robot reaches the end of the last segment.

\begin{algorithm}[t]
\caption{Observability-Aware Active Calibration}\label{alg:3-1} 
\begin{algorithmic} 
{\small
\STATE\textbf{Input:} Sensor measurements; \\
\STATE\textbf{Output:} Estimation of sensor extrinsic parameters $\hat{\mathbf{\psi}}$; \\
\STATE\textbf{Parameter:} Sound source position $\mathbf{s}$, initial guesses $\hat{\mathbf{\psi}}_0$, number of control points $n$, motion boundary $\eta$, steering threshold $\beta$, iteration threshold $\zeta$.

~\\

\STATE Initialize calibration trajectory $\mathbf{T}_{1:n-3}$;

\WHILE{$\hat{\mathbf{\psi}}$ not converged \textbf{and} $j\leq n-3$}
    \STATE Robot moves along the trajectory segment $\mathbf{T}_{j}$; 

    \STATE Collect sensor measurements and update the extrinsic parameters with (\ref{predict})-(\ref{update}) in real-time;

    \IF{LiDAR extrinsic parameter $\hat{\mathbf{\psi}}_L$ converged}
    \STATE Estimate the robot pose $^{G}\mathbf{t}_{k}$ and $^{G}\theta_{k}$ by LiDAR odometry for microphone array calibration;
    \ENDIF
    
    \IF{Robot arrived at the end of the segment $\mathbf{T}_{j}$}
    \STATE Optimize the subsequent trajectory $\mathbf{T}_{j+1:n-3}$ via ($\ref{opt_Q}$);
    \ENDIF

\ENDWHILE
}

\end{algorithmic} 
\end{algorithm}

\subsection{\label{pre-1-1}Online Calibration for Multimodal Sensor System}

To perform online calibration under limited computational capability, we employs the widely used EKF to recursively update the sensor state parameters in real-time. Denote $\hat{\psi}$ and $\hat{\mathbf{P}}$ as the estimated mean and the variance of sensor extrinsic parameters, respectively. During calibration, the sensor state parameters are assumed to evolve statically. The prediction and correction steps are
\begin{equation}\label{predict}
\mathbf{\hat{\psi}}_{k|k-1}=\mathbf{\hat{\psi}}_{k-1|k-1},\text{ }\mathbf{\hat{P}}_{k|k-1}=\hat{\mathbf{P}}_{k-1|k-1}
\end{equation}
and 
\begin{equation}\label{update}
\begin{aligned}\hat{\mathbf{\psi}}_{k|k}= & \mathbf{\hat{\psi}}_{k|k-1}+\mathbf{K}_{k}\left(\mathbf{z}_{k}-\mathbf{g}(^{G}\mathbf{t}_{k},^{G}\theta_{k},\hat{\psi}_{k|k-1})\right)\\
\mathbf{\hat{P}}_{k|k}= & \left(\mathbf{I}-\mathbf{K}_{k}\mathbf{G}_k\right)\mathbf{\hat{P}}_{k|k-1}
\end{aligned},
\end{equation}
respectively, where $\mathbf{K}_{k}$ is called the Kalman gain and is calculated as
\begin{equation}
\mathbf{K}_{k}=\mathbf{\hat{P}}_{k|k-1}\mathbf{G}_k^{\top}\left(\mathbf{G}_k\mathbf{\hat{P}}_{k|k-1}\mathbf{G}_k^{\top}+\mathbf{N}_{k}\right)^{-1},
\end{equation}
where $\mathbf{G}_k$ is the Jacobian matrix of the observation model in (\ref{eq:system}), and can be calculated as
\begin{equation}
\mathbf{G}_k=\left(\frac{\partial\mathbf{g}}{\mathbf{\partial\psi}}\right)\vert_{^{G}\mathbf{t}_{k},^{G}\theta_{k},\mathbf{\hat{\psi}}_{k|k-1}}\\
= \begin{bmatrix}
\begin{smallmatrix}
\frac{\partial\mathbf{d}_{k}}{\psi_{M}} & \mathbf{0}\\
\mathbf{0} & \frac{\partial\Delta\mathbf{t}_{L,k}}{\psi_{L}}\\
\mathbf{0} & \mathbf{0}
\end{smallmatrix}
\end{bmatrix}\vert_{^{G}\mathbf{t}_{k},^{G}\theta_{k},\mathbf{\hat{\psi}}_{k|k-1}}.
\end{equation}


\begin{figure*}[htbp]
	\centering
	\begin{minipage}{0.32\linewidth}
		\centering
		\subfigure[]{\includegraphics[width=0.8\linewidth]{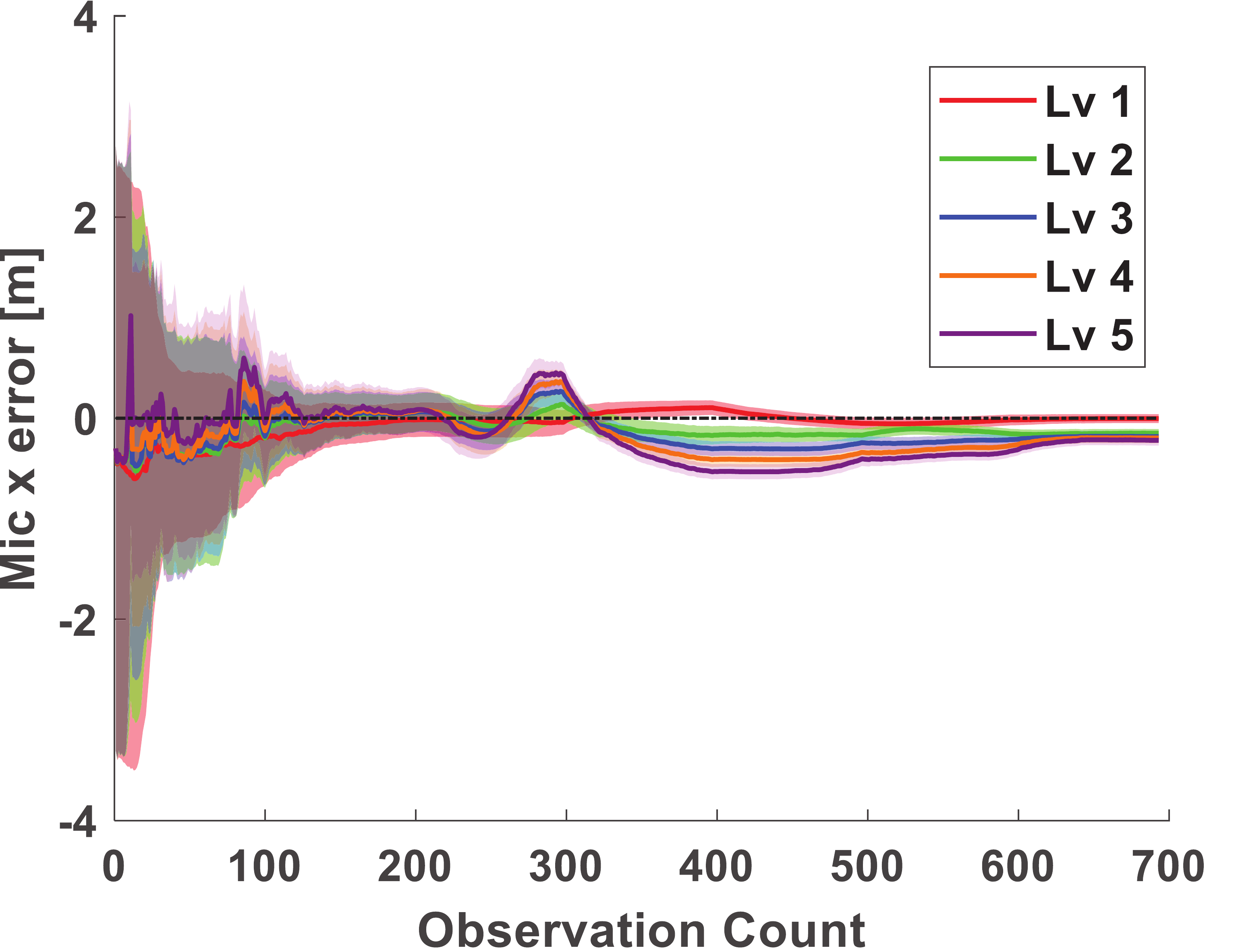}}
	\end{minipage}
	\begin{minipage}{0.32\linewidth}
		\centering
		\subfigure[]{\includegraphics[width=0.8\linewidth]{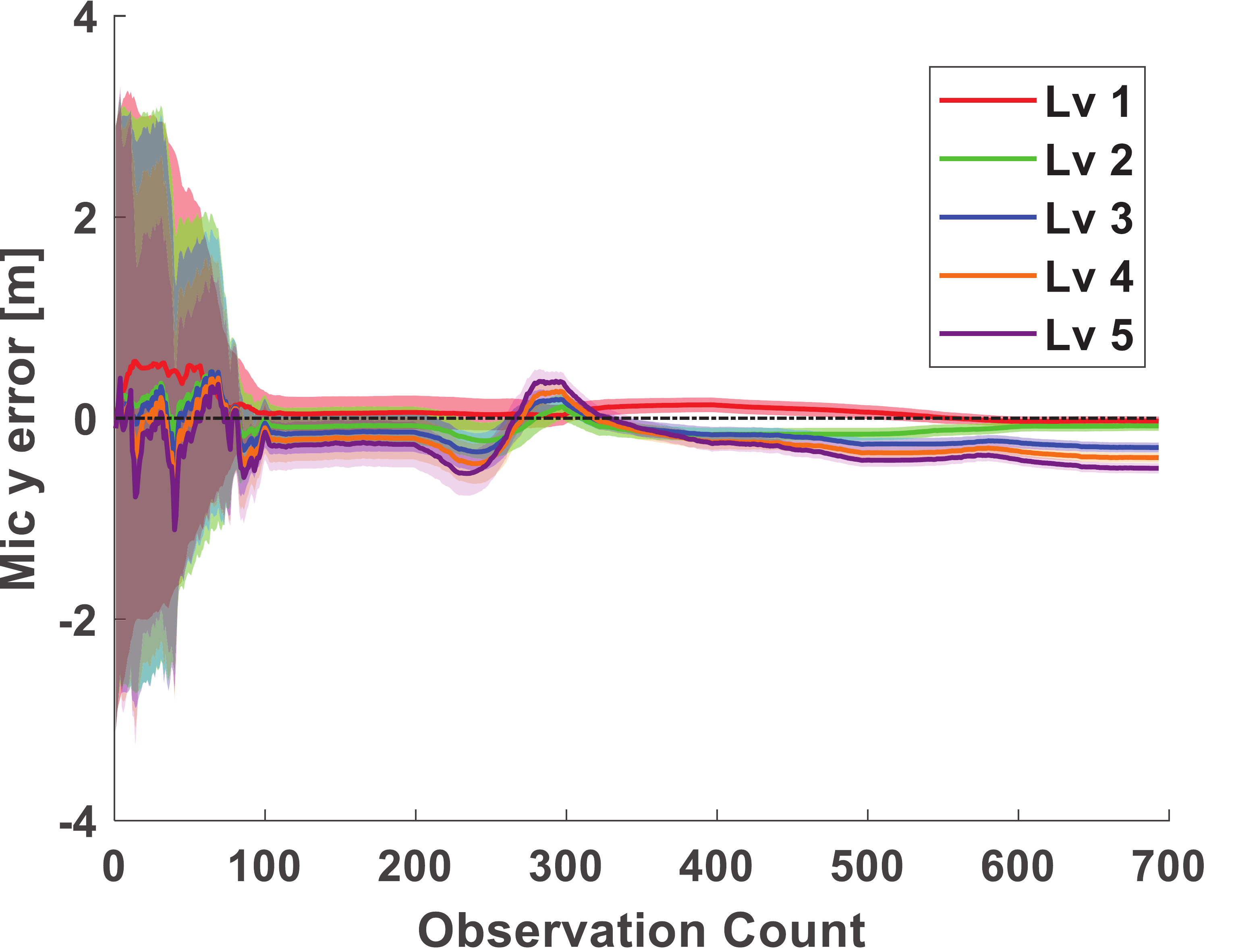}}
	\end{minipage}
	\begin{minipage}{0.32\linewidth}
		\centering
		\subfigure[]{\includegraphics[width=0.8\linewidth]{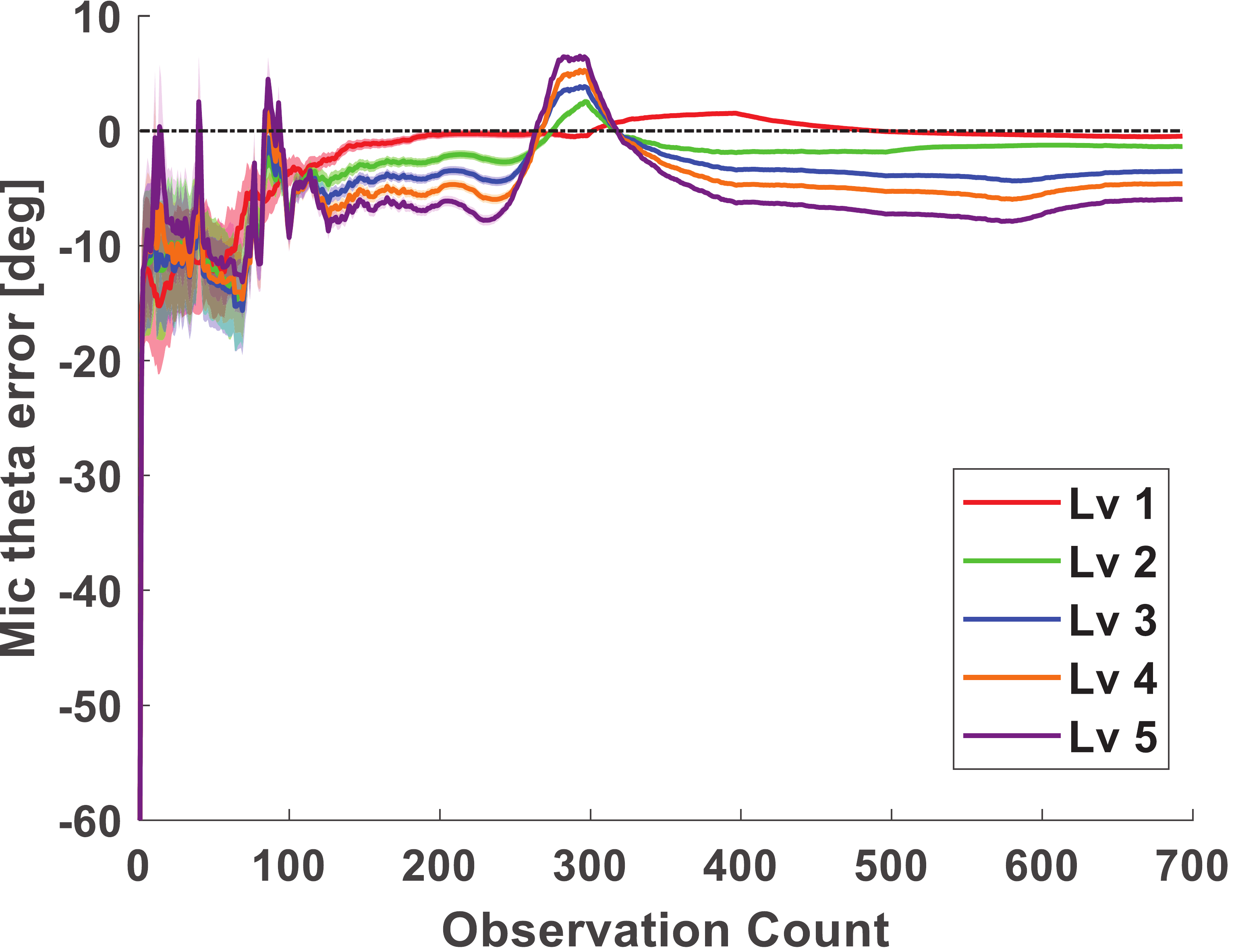}}
	\end{minipage}
    \qquad
	\begin{minipage}{0.32\linewidth}
		\centering
		\subfigure[]{\includegraphics[width=0.8\linewidth]{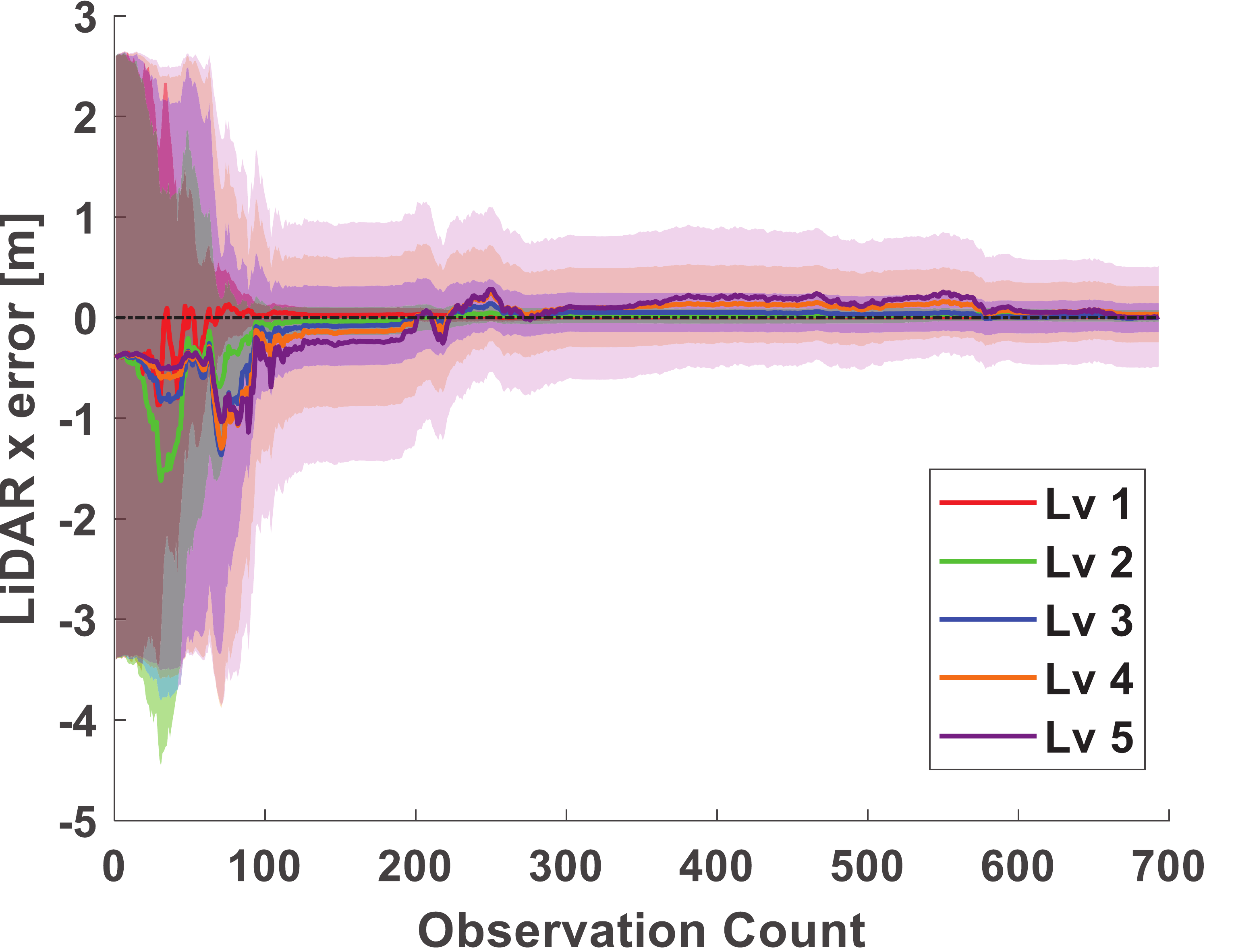}}
	\end{minipage}
	\begin{minipage}{0.32\linewidth}
		\centering
		\subfigure[]{\includegraphics[width=0.8\linewidth]{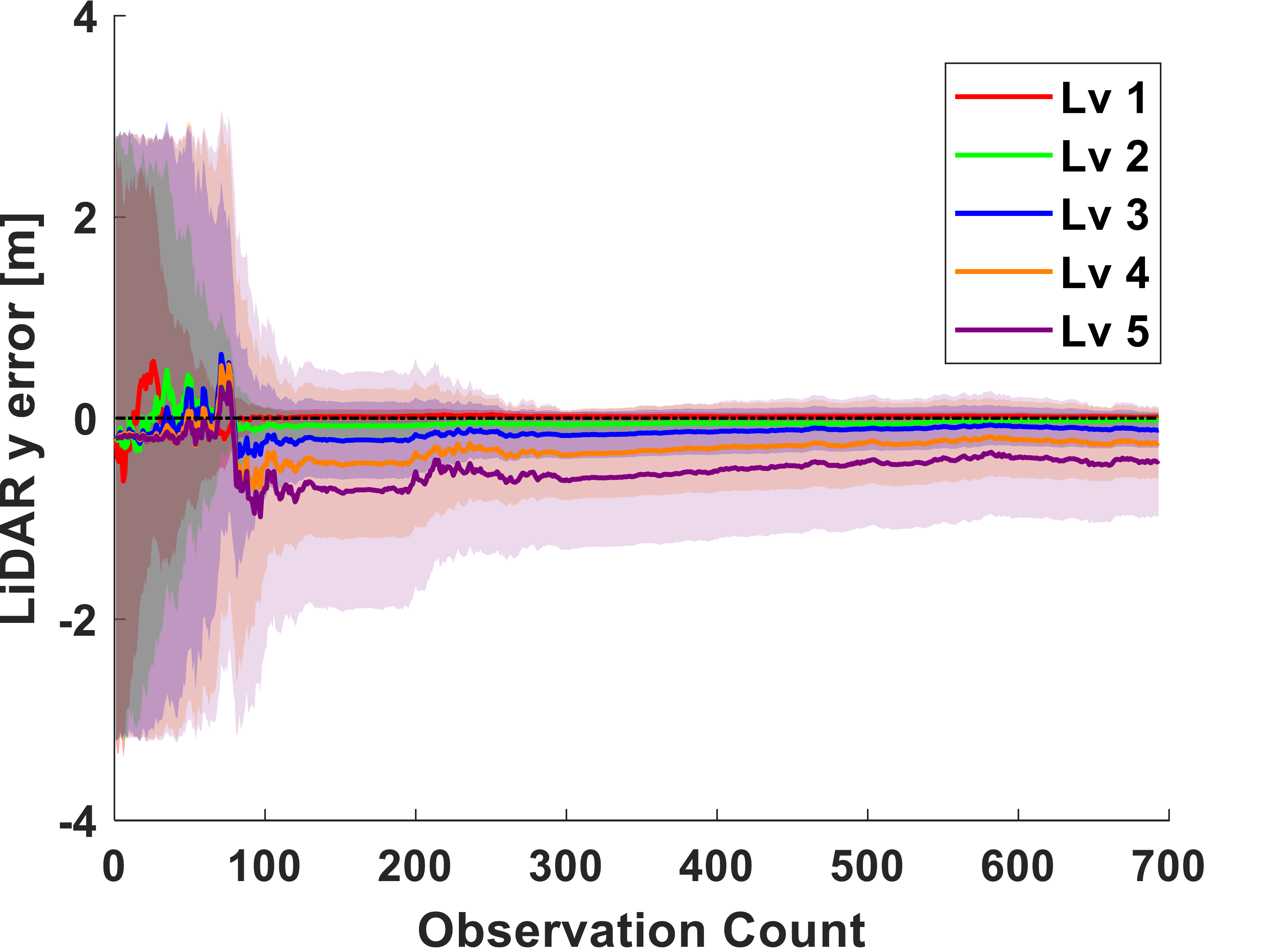}}
	\end{minipage}
    \begin{minipage}{0.32\linewidth}
		\centering
		\subfigure[]{\includegraphics[width=0.8\linewidth]{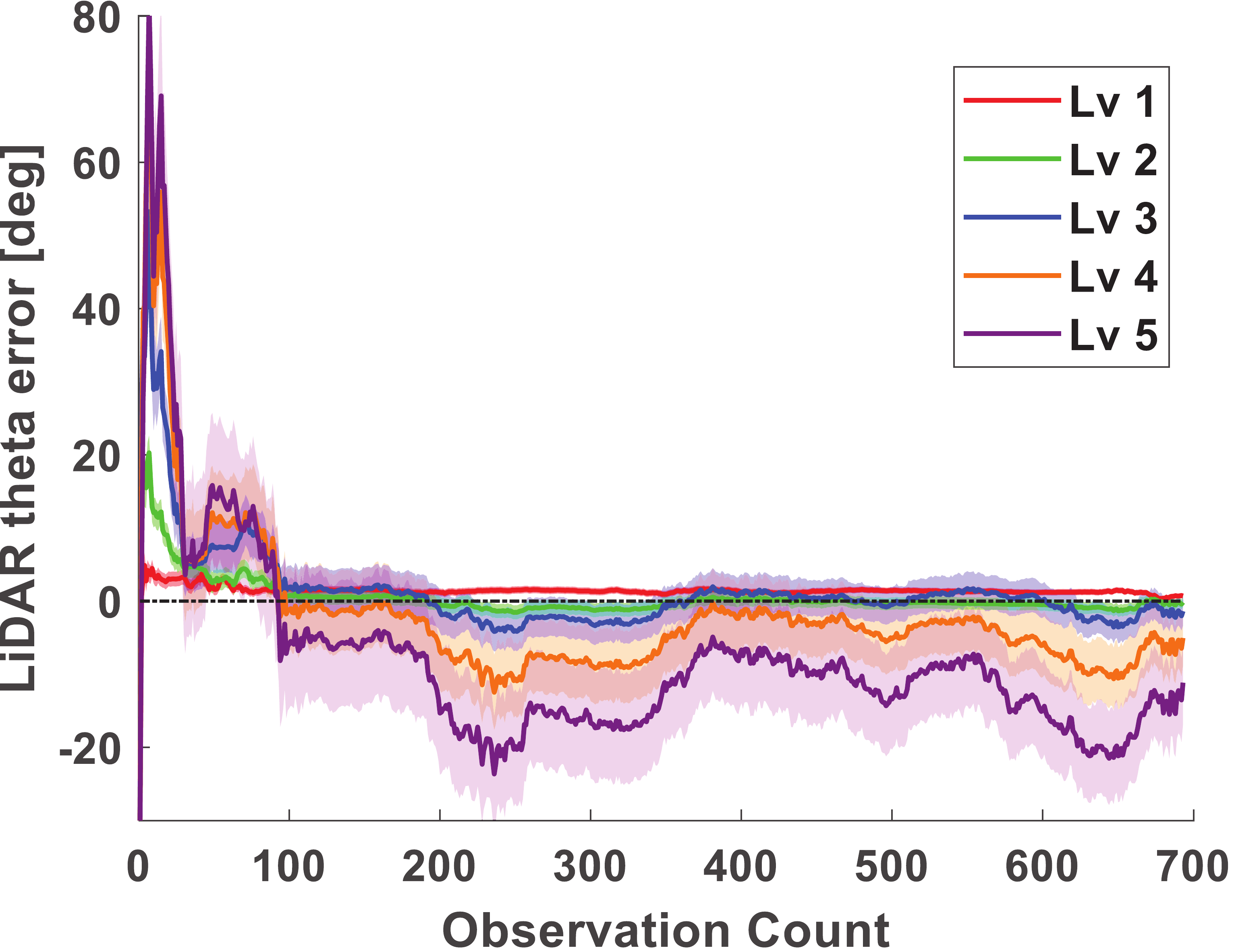}}
	\end{minipage}
	\caption{Calibration errors under varying levels of measurement noise. The solid lines represent the estimation errors of the calibration results in simulation, while the shaded regions denote the corresponding 3$\sigma$ bounds.} 
    \label{fig:3}
\end{figure*}
\begin{figure*}[htbp]
	\centering
	\begin{minipage}{0.32\linewidth}
		\centering
		\subfigure[]{\includegraphics[width=0.8\linewidth]{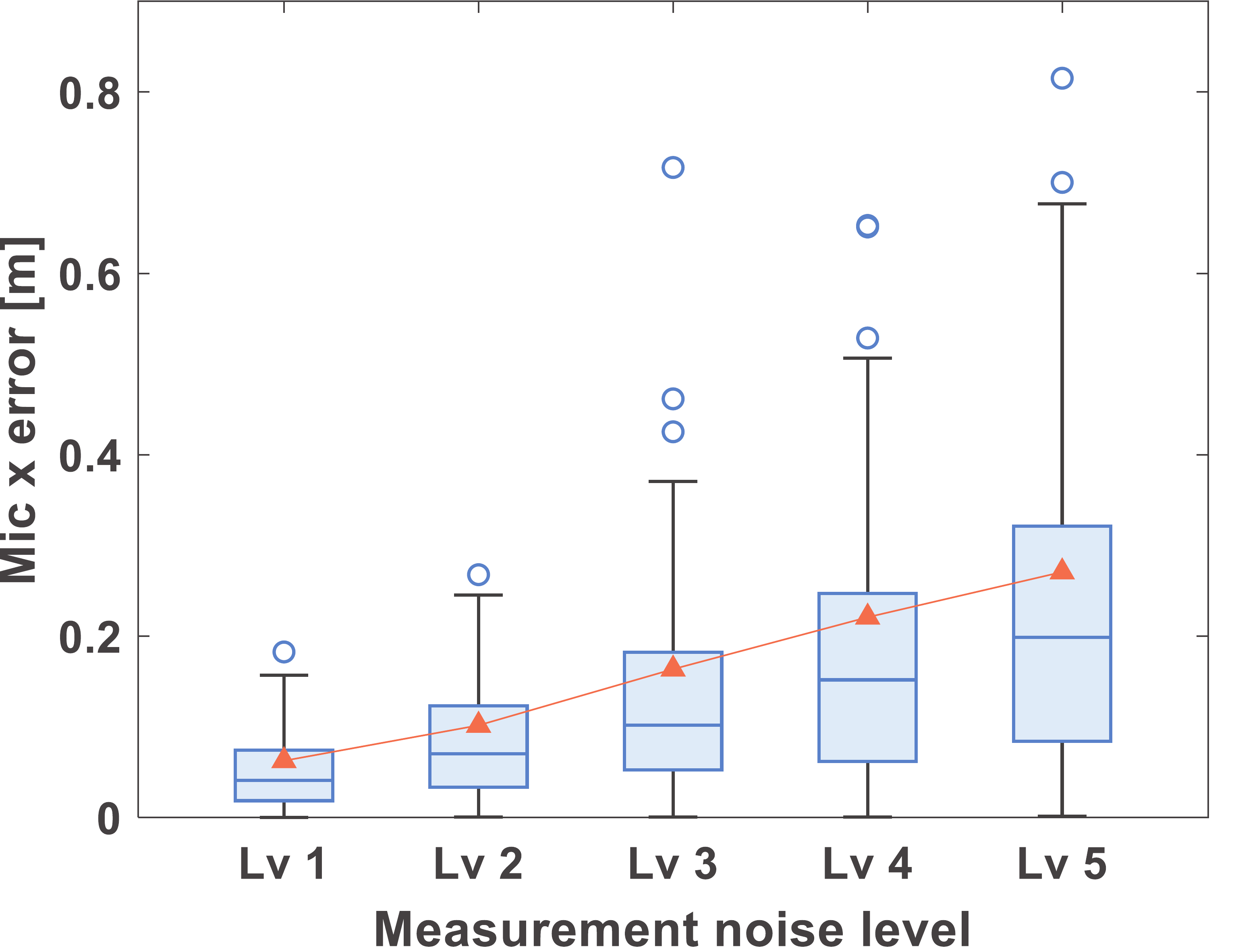}}
	\end{minipage}
	\begin{minipage}{0.32\linewidth}
		\centering
		\subfigure[]{\includegraphics[width=0.8\linewidth]{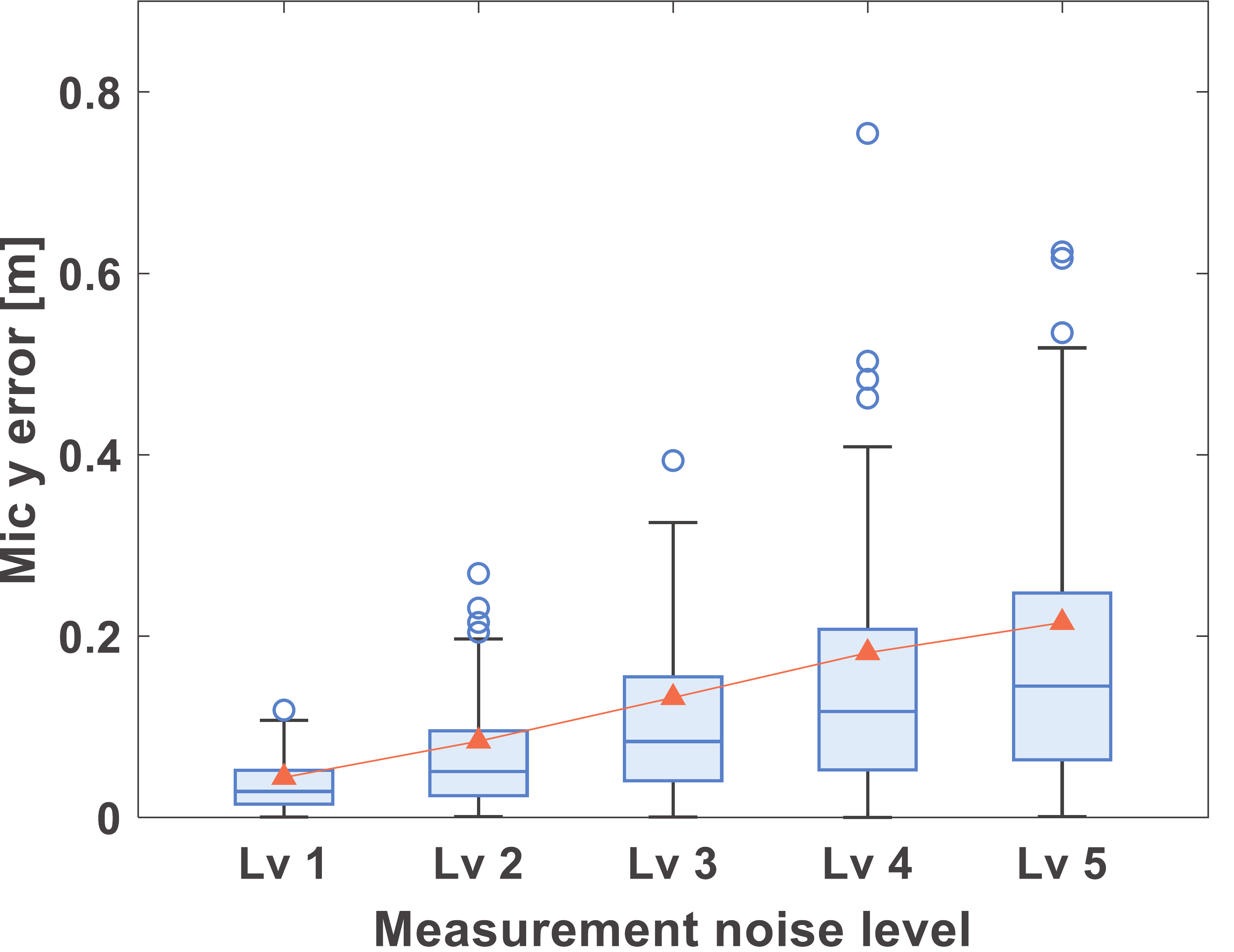}}
	\end{minipage}
	\begin{minipage}{0.32\linewidth}
		\centering
		\subfigure[]{\includegraphics[width=0.8\linewidth]{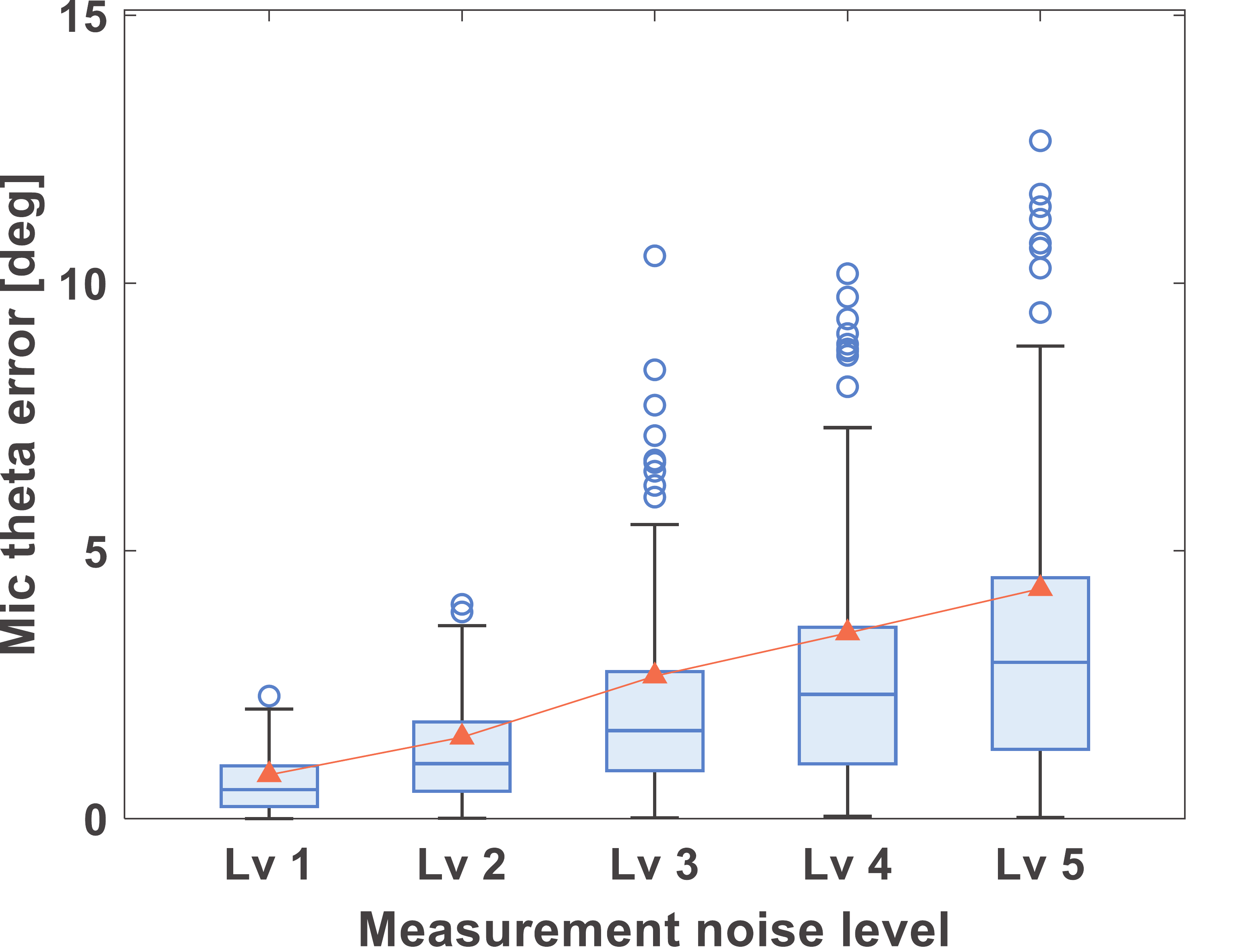}}
	\end{minipage}
	\qquad
	\begin{minipage}{0.32\linewidth}
		\centering
		\subfigure[]{\includegraphics[width=0.8\linewidth]{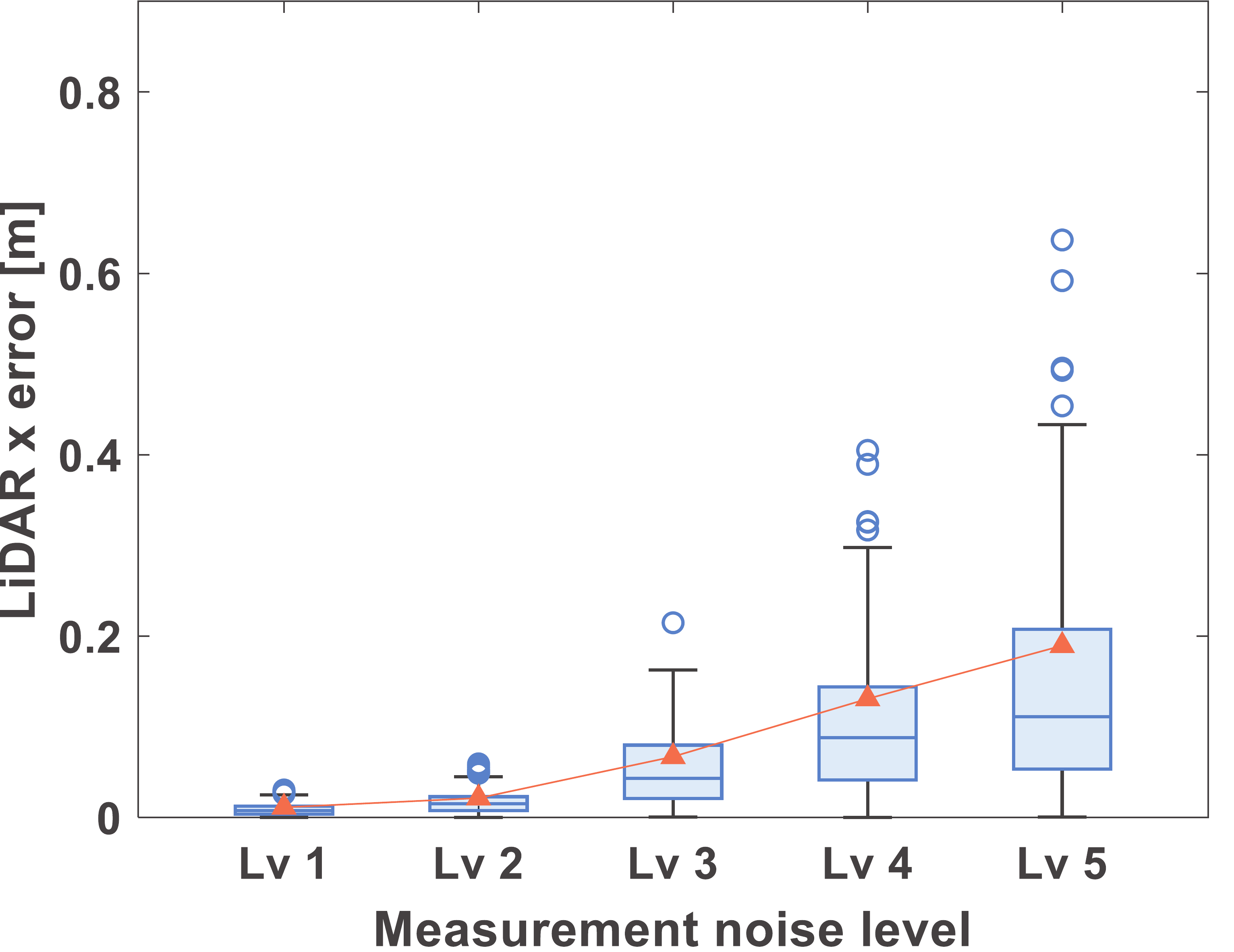}}
	\end{minipage}
	\begin{minipage}{0.32\linewidth}
		\centering
		\subfigure[]{\includegraphics[width=0.8\linewidth]{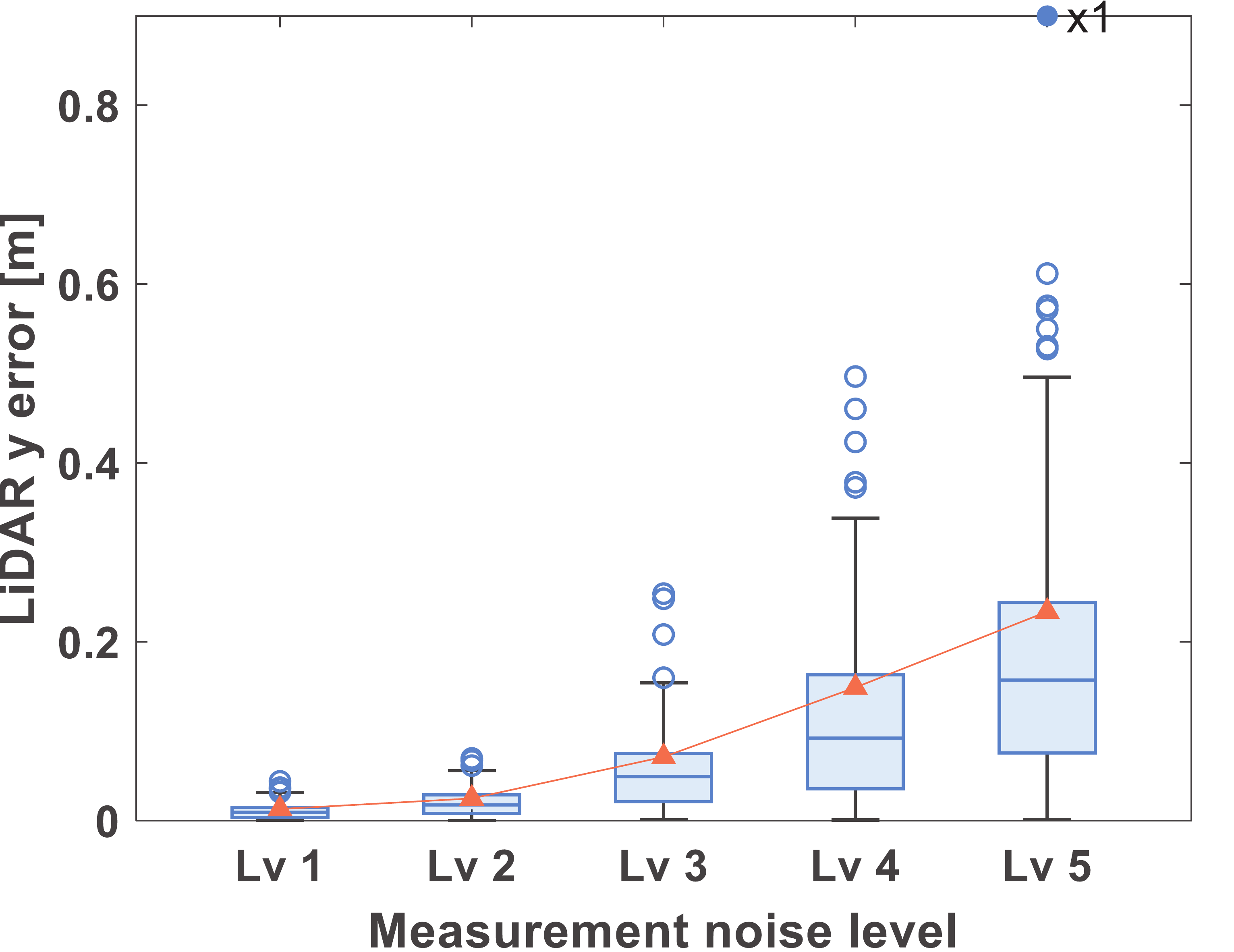}}
	\end{minipage}
    \begin{minipage}{0.32\linewidth}
		\centering
		\subfigure[]{\includegraphics[width=0.8\linewidth]{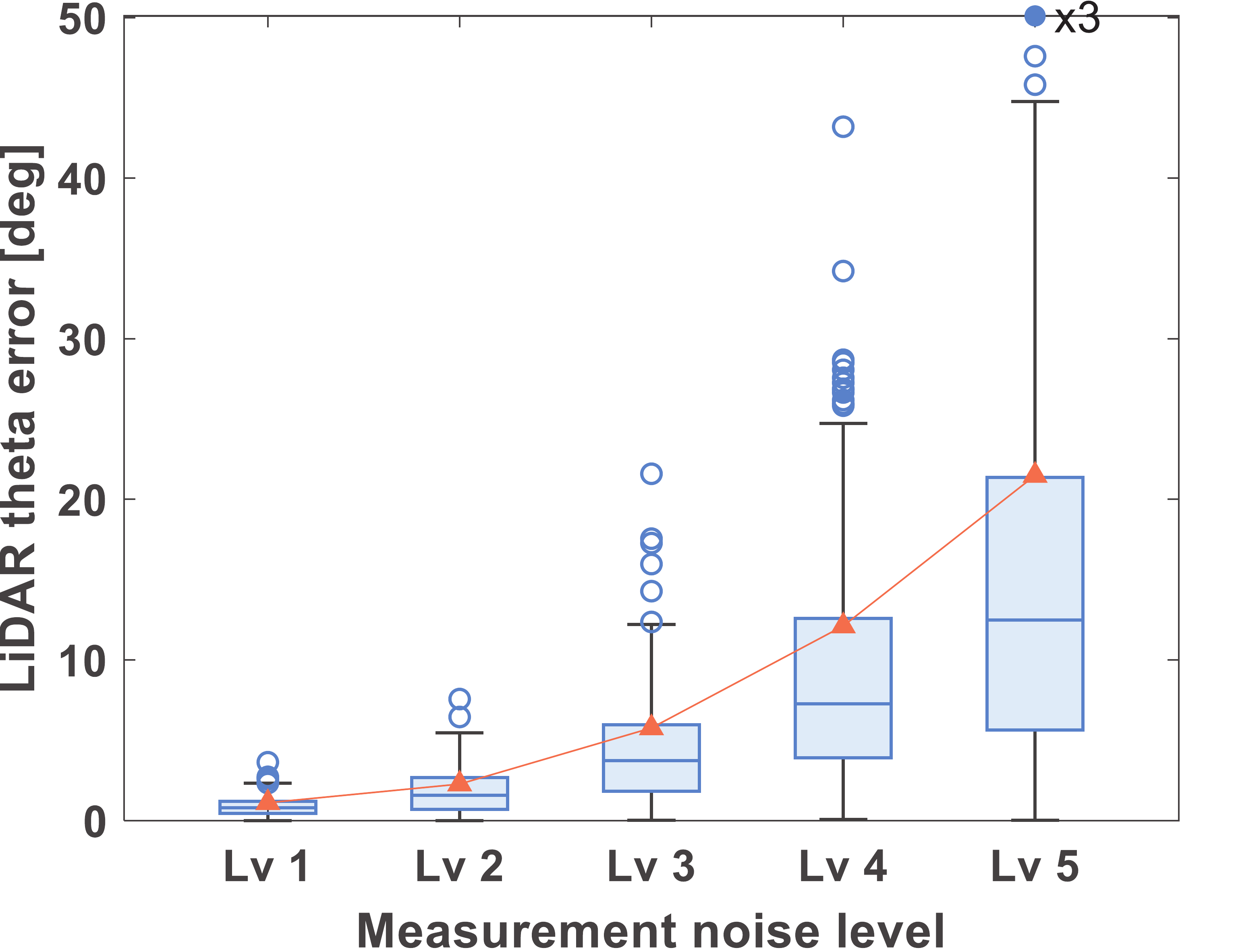}}
	\end{minipage}
	\caption{Error distributions between the calibration results and true values for 200 Monte Carlo simulations under varying levels of measurement noise.} 
    \label{fig:4}
\end{figure*}
\begin{figure*}[htbp]
	\centering
	\begin{minipage}{0.32\linewidth}
		\centering
		\subfigure[]{\includegraphics[width=0.8\linewidth]{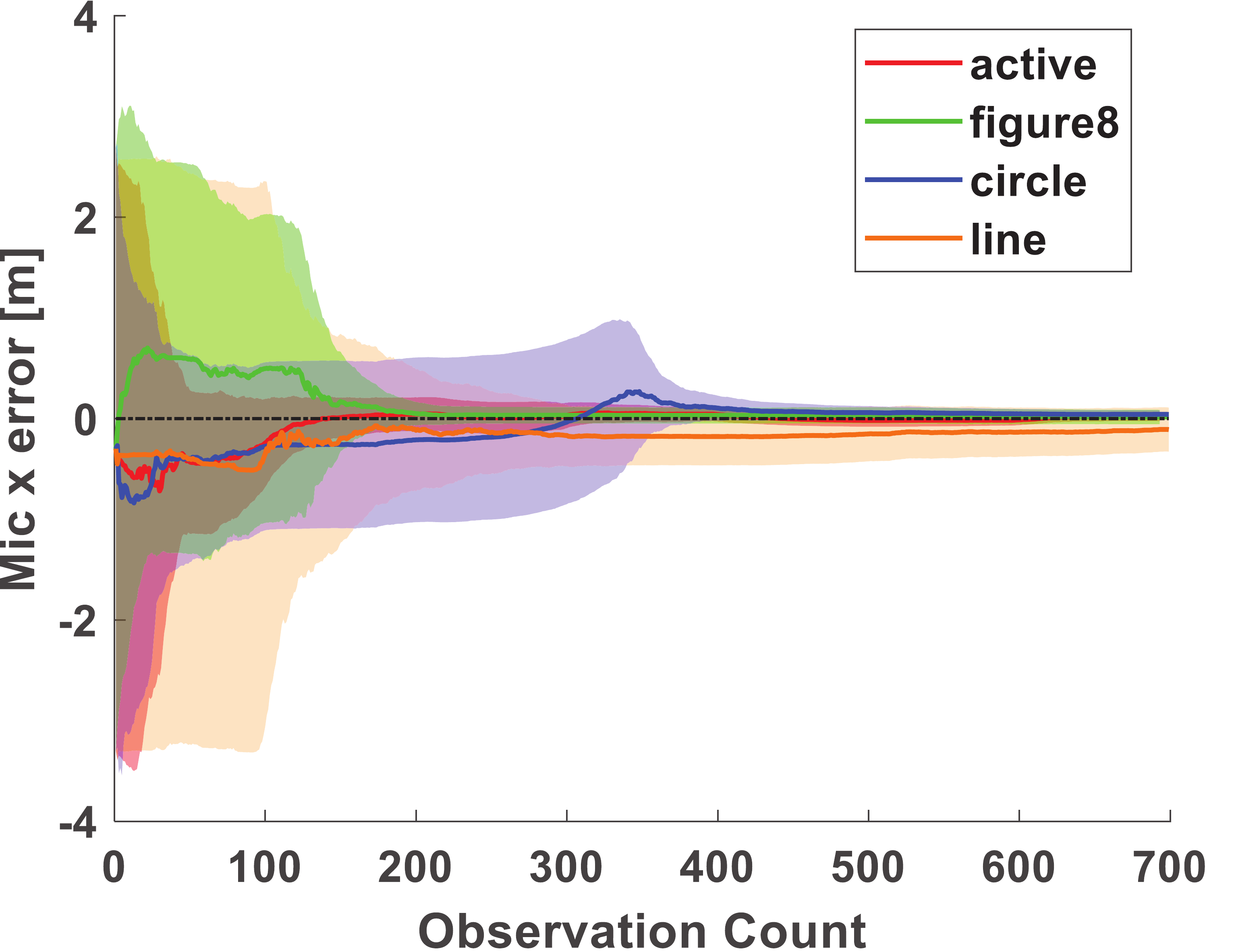}}
	\end{minipage}
	\begin{minipage}{0.32\linewidth}
		\centering
		\subfigure[]{\includegraphics[width=0.8\linewidth]{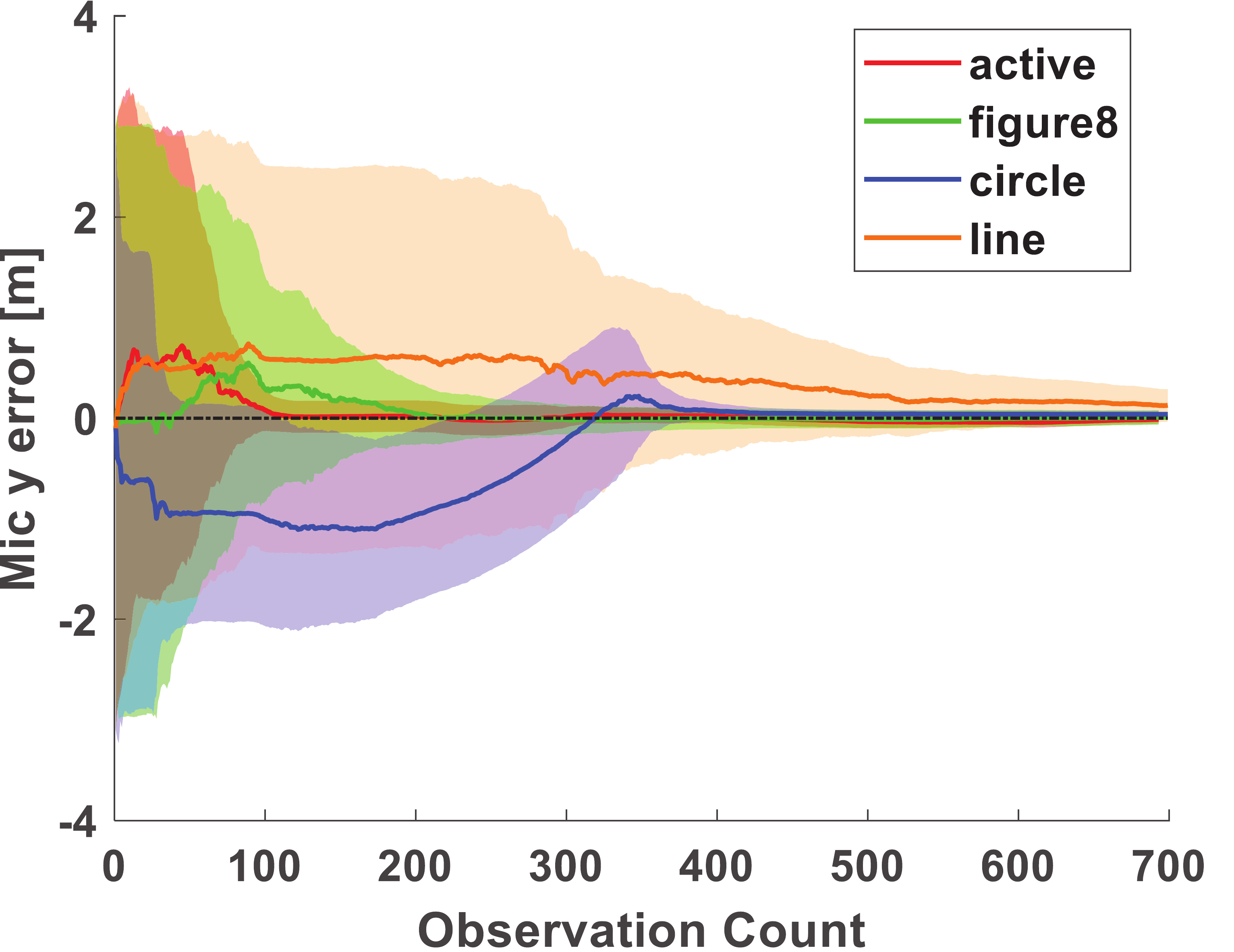}}
	\end{minipage}
	\begin{minipage}{0.32\linewidth}
		\centering
		\subfigure[]{\includegraphics[width=0.8\linewidth]{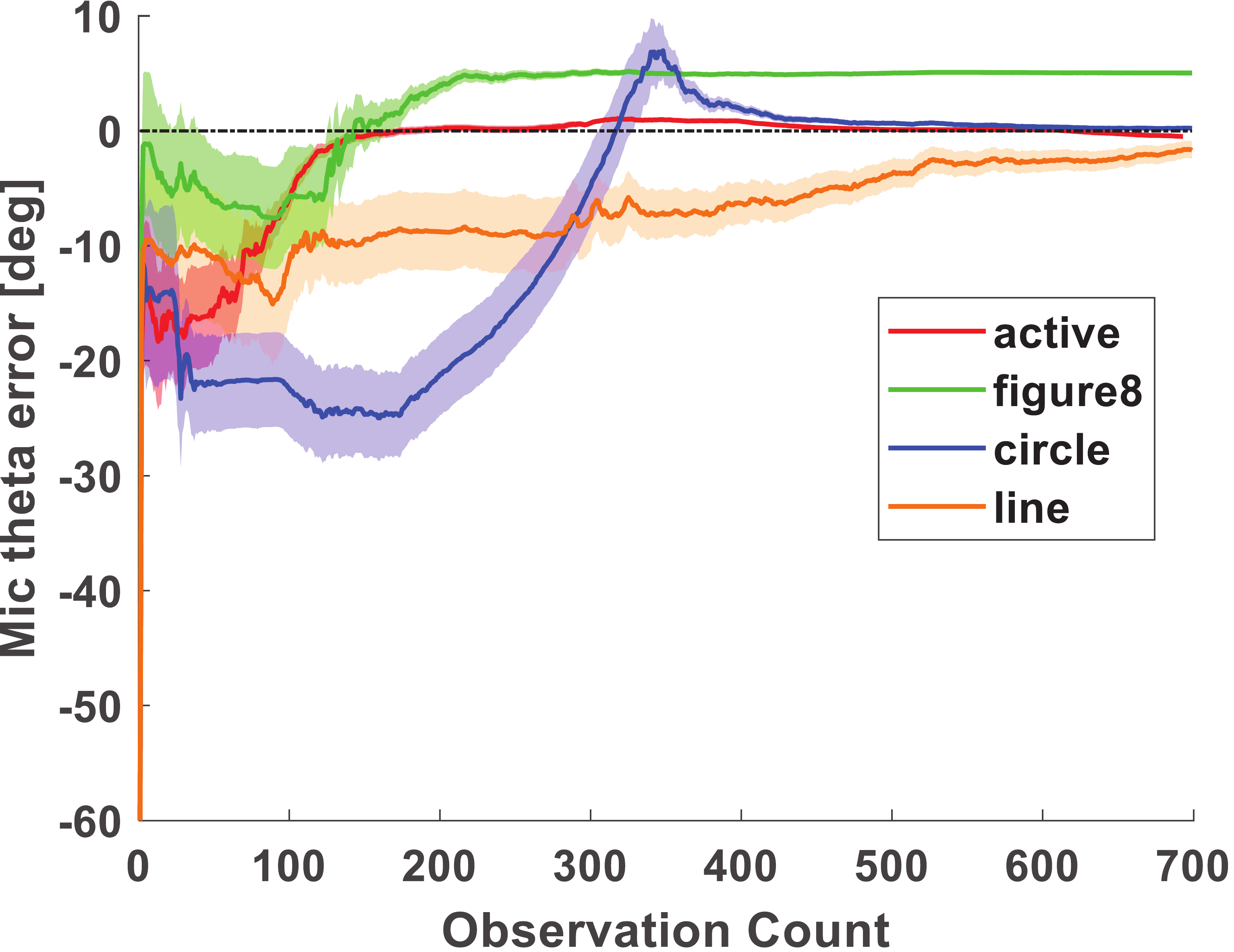}}
	\end{minipage}
    \qquad
	\begin{minipage}{0.32\linewidth}
		\centering
		\subfigure[]{\includegraphics[width=0.8\linewidth]{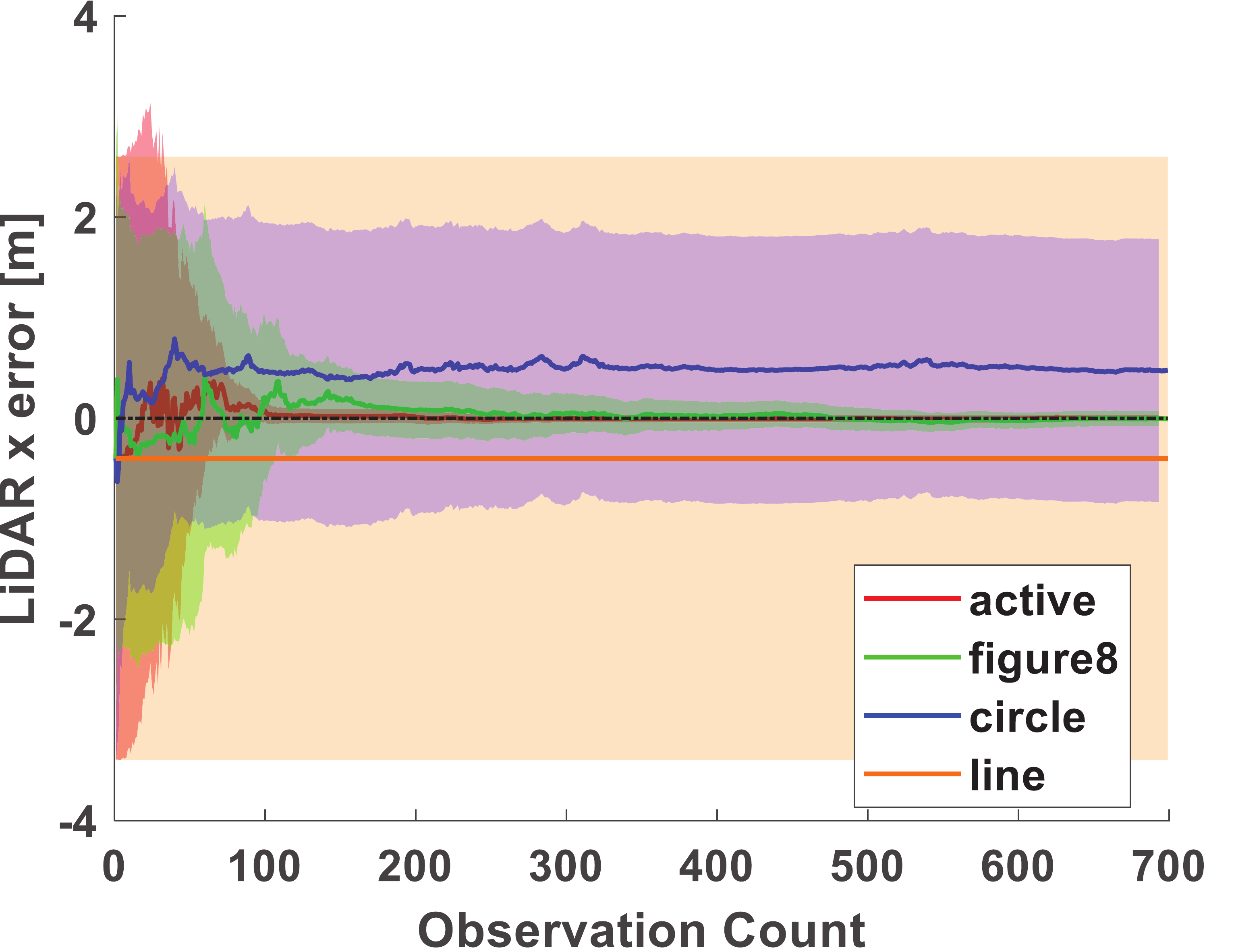}}
	\end{minipage}
	\begin{minipage}{0.32\linewidth}
		\centering
		\subfigure[]{\includegraphics[width=0.8\linewidth]{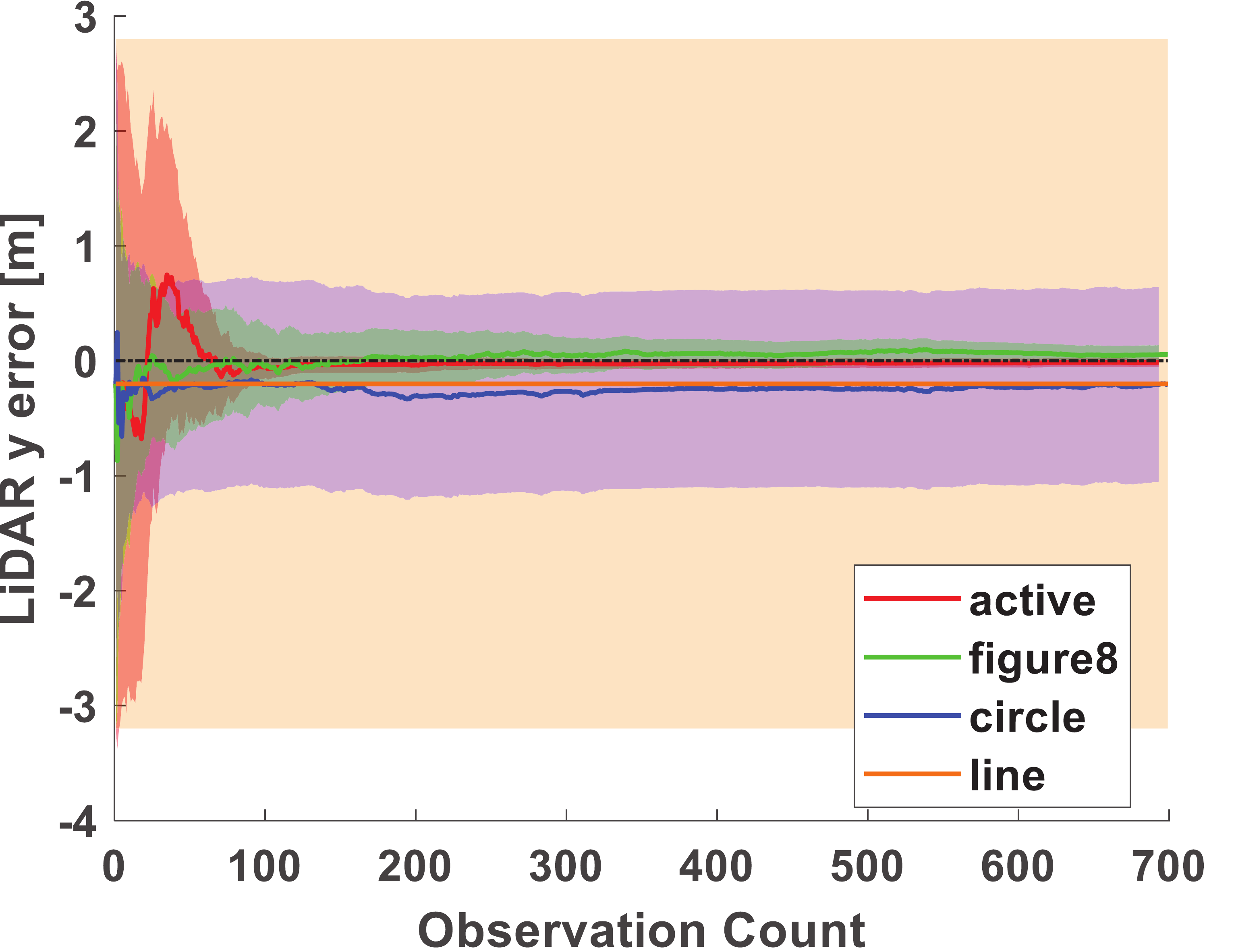}}
	\end{minipage}
    \begin{minipage}{0.32\linewidth}
		\centering
		\subfigure[]{\includegraphics[width=0.8\linewidth]{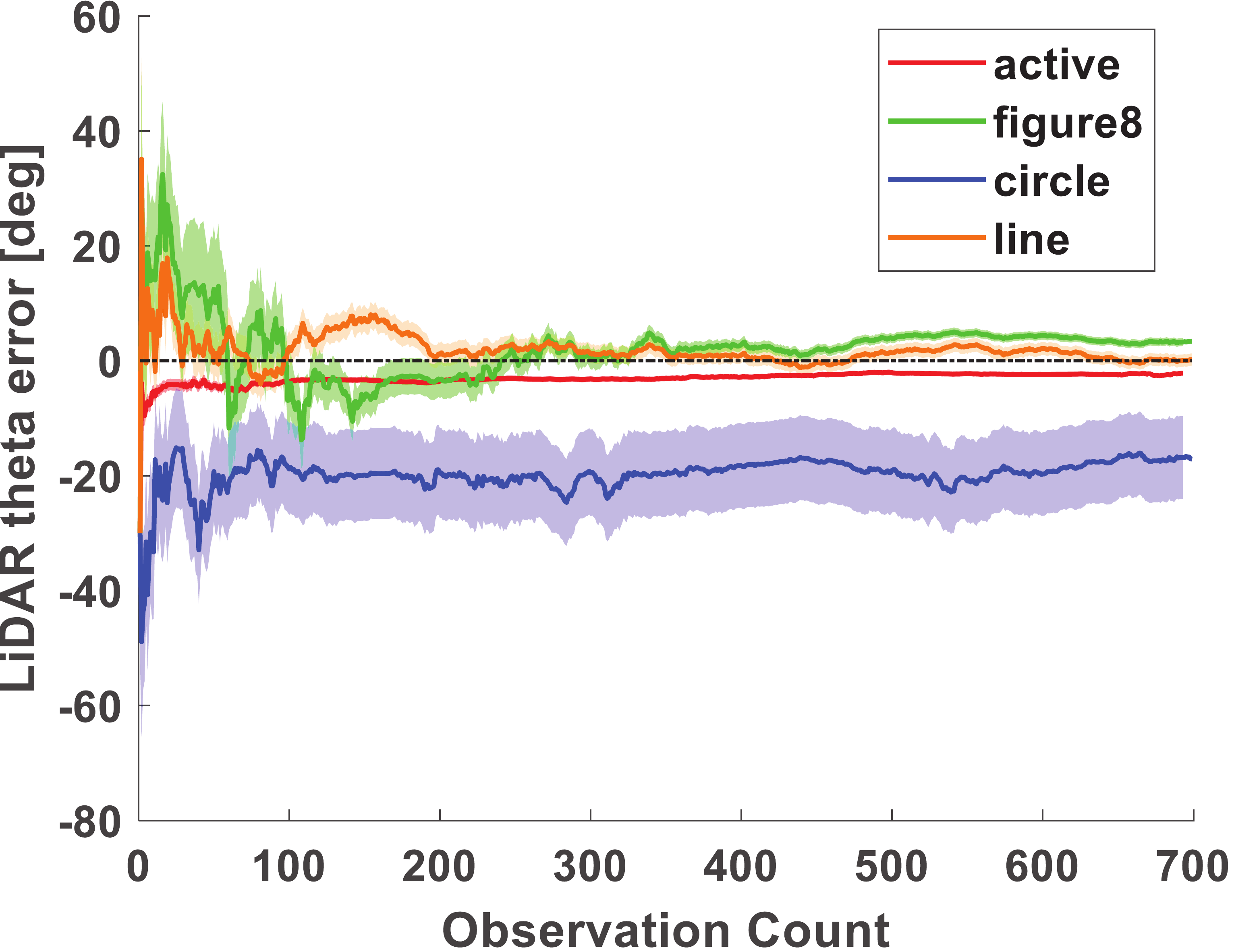}}
	\end{minipage}
	\caption{The solid lines represent the estimation errors of the calibration results (obtained from commonly employed fixed trajectories and the proposed active calibration method), while the shaded regions denote the corresponding 3$\sigma$ bounds.} 
    \label{fig:5}
\end{figure*}
\begin{figure*}[htbp]
	\centering
	\begin{minipage}{0.32\linewidth}
		\centering
		\subfigure[]{\includegraphics[width=0.8\linewidth]{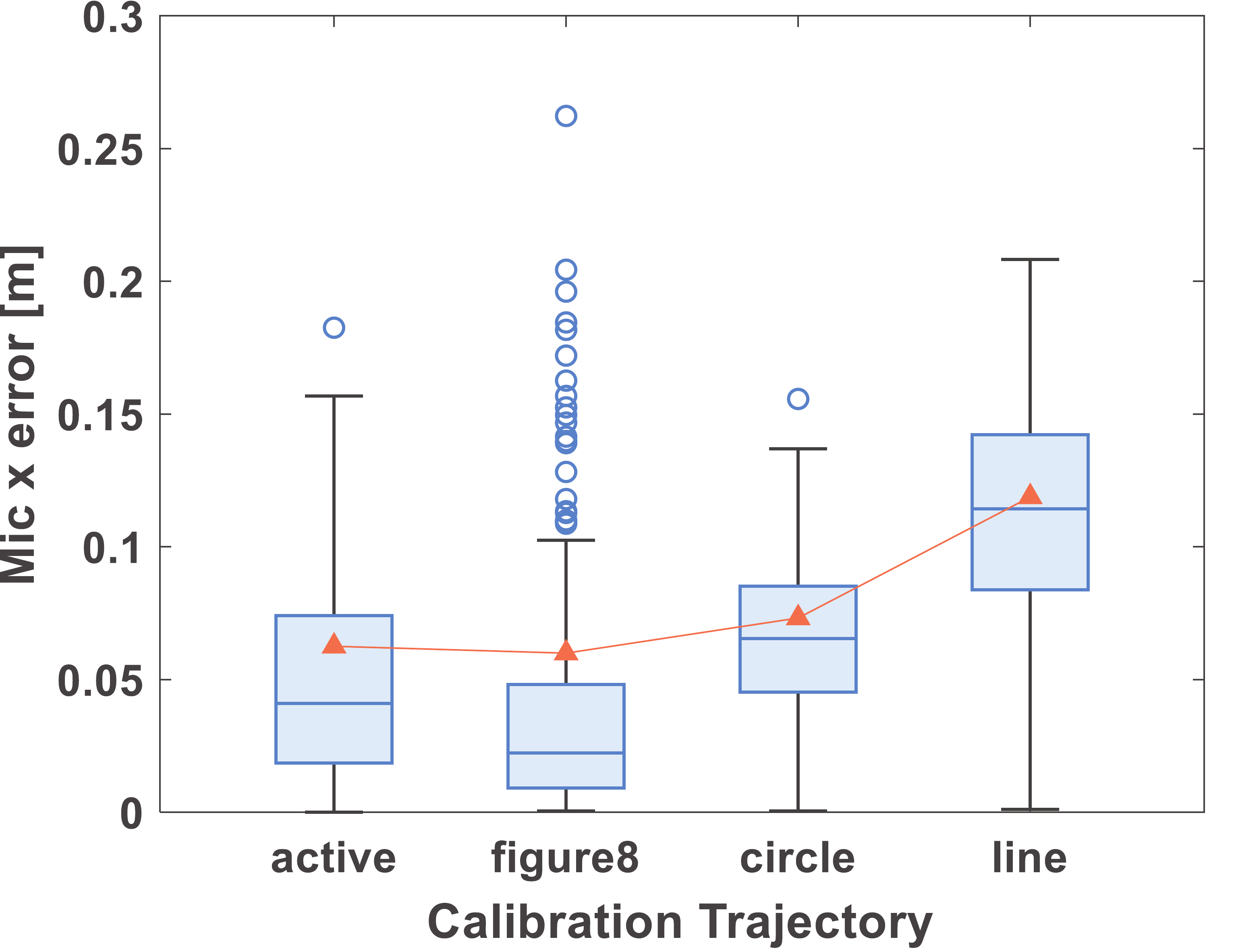}}
	\end{minipage}
	\begin{minipage}{0.32\linewidth}
		\centering
		\subfigure[]{\includegraphics[width=0.8\linewidth]{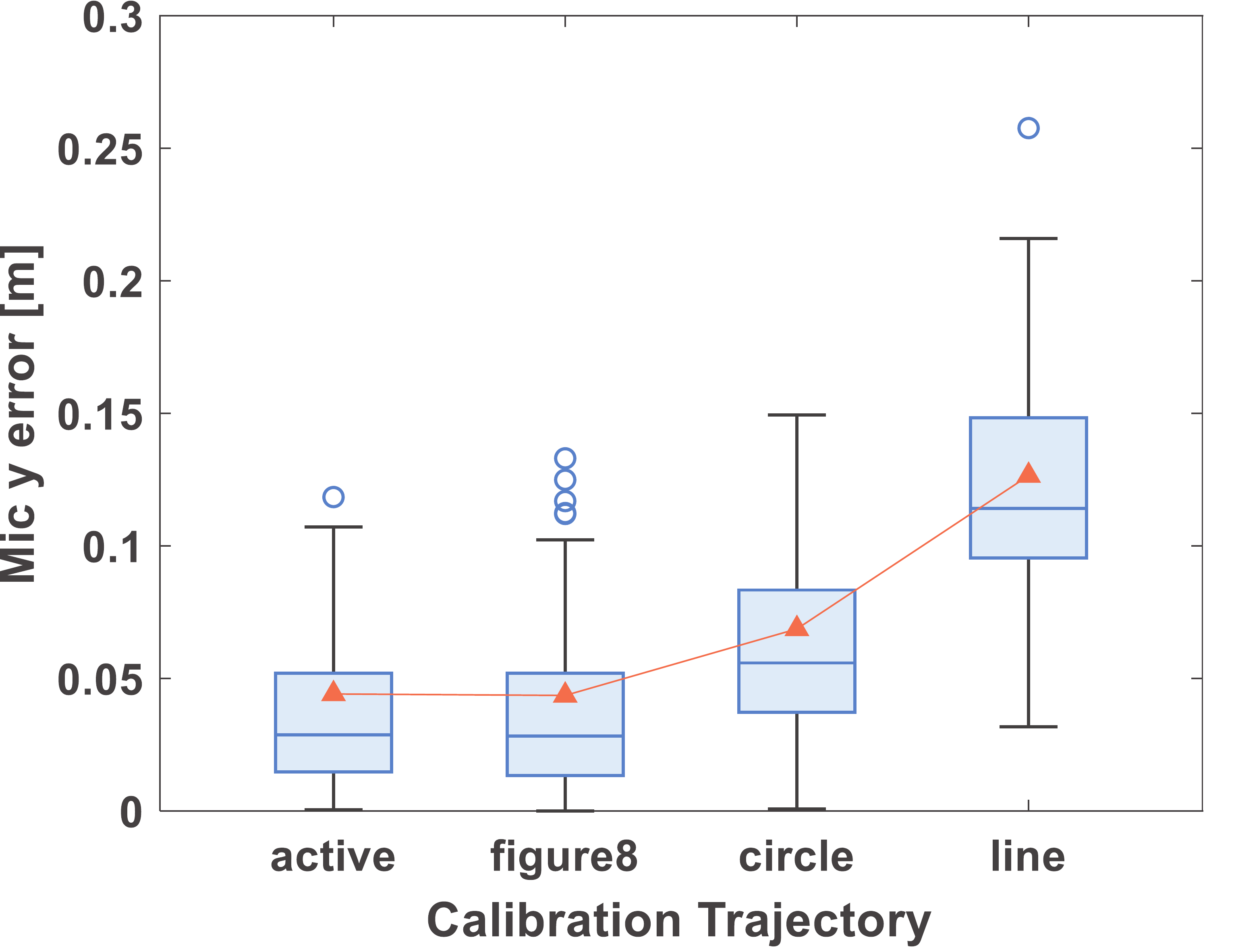}}
	\end{minipage}
	\begin{minipage}{0.32\linewidth}
		\centering
		\subfigure[]{\includegraphics[width=0.8\linewidth]{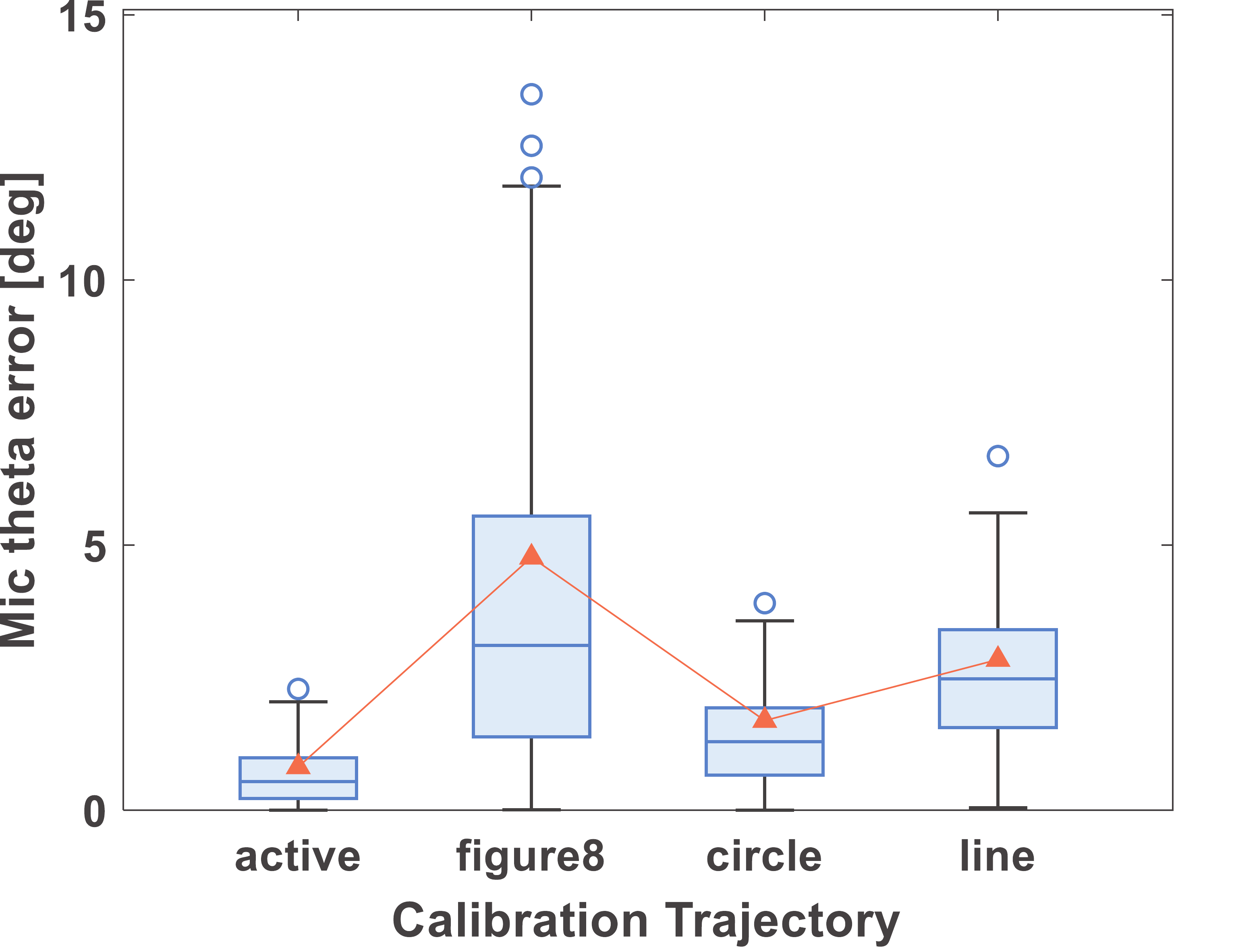}}
	\end{minipage}
    \qquad
	\begin{minipage}{0.32\linewidth}
		\centering
		\subfigure[]{\includegraphics[width=0.8\linewidth]{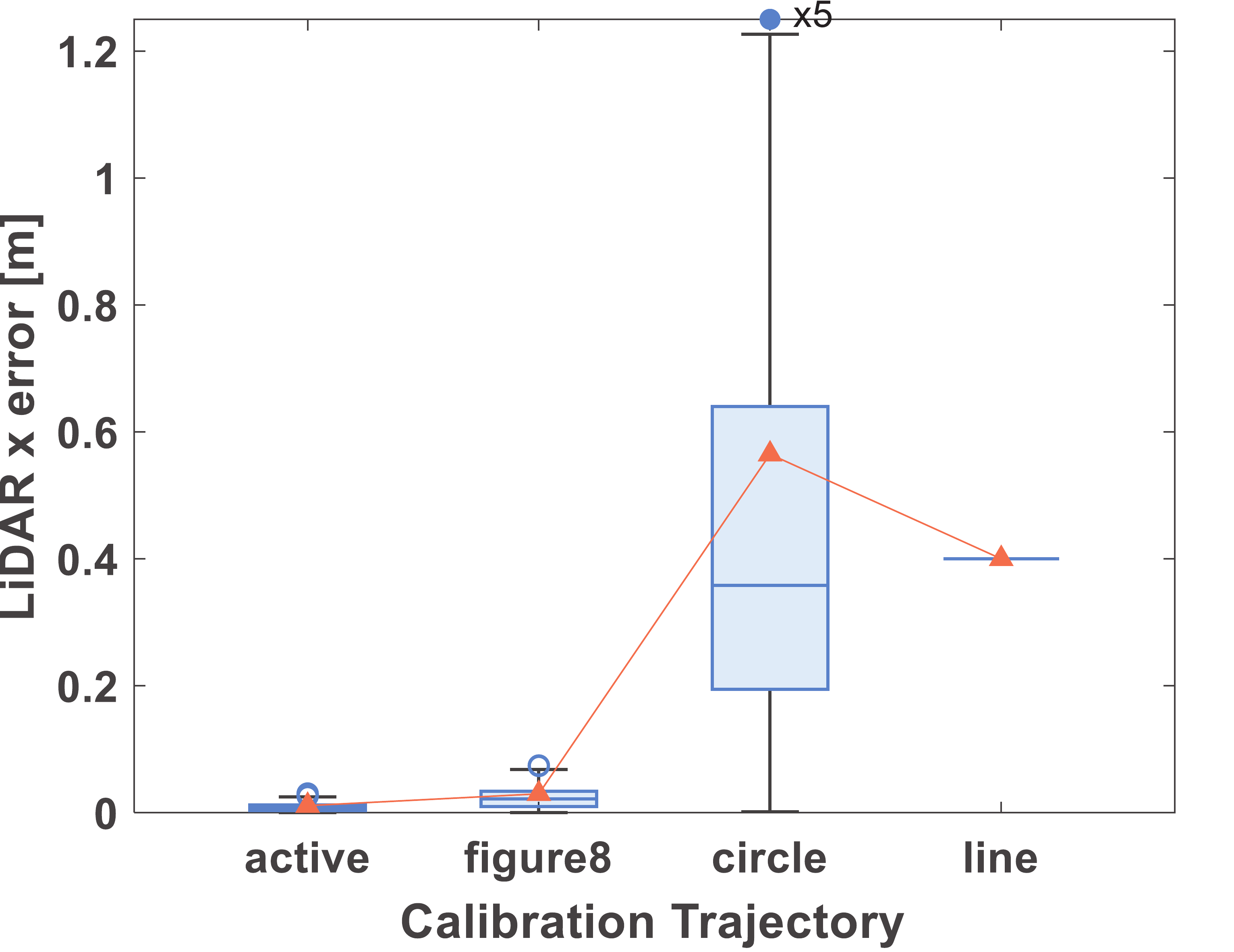}}
	\end{minipage}
	\begin{minipage}{0.32\linewidth}
		\centering
		\subfigure[]{\includegraphics[width=0.8\linewidth]{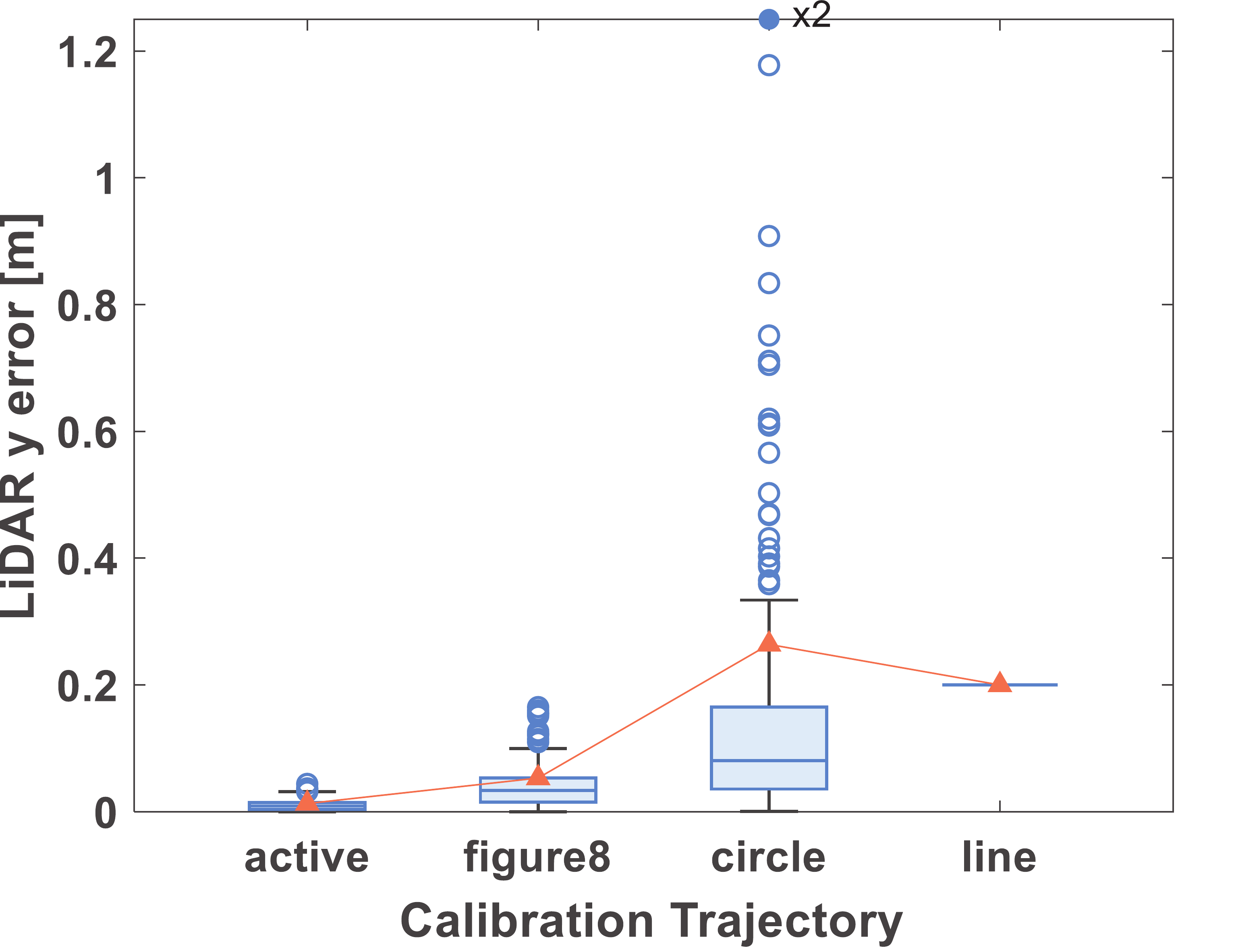}}
	\end{minipage}
    \begin{minipage}{0.32\linewidth}
		\centering
		\subfigure[]{\includegraphics[width=0.8\linewidth]{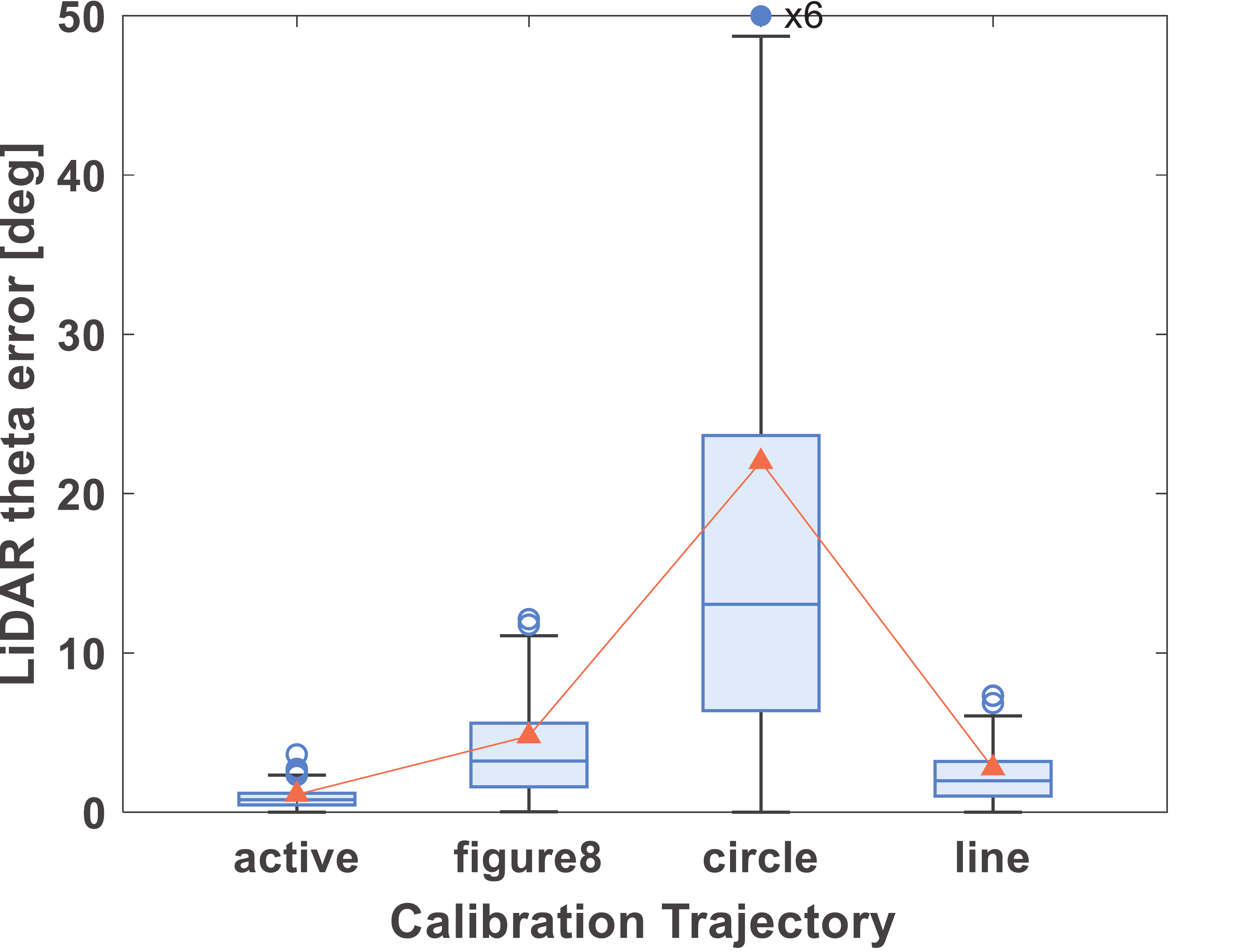}}
	\end{minipage}
	\caption{Error distributions between the calibration results and true values for 200 Monte Carlo simulations under different trajectory strategies.} 
    \label{fig:6}
\end{figure*}


\subsection{Potential Applications for 3D Scenario and Other Sensor Modalities}
Compared to the 2D case, active calibration of sensor extrinsics in 3D space inevitably introduces new challenges, including an increased number of calibration parameters (6 degrees of freedom for each sensor pose), more sophisticated sensor measurements, and greater demands on motion planning and control in 3D space. Although the proposed calibration framework is developed and validated in a two-dimensional environment involving a microphone array, wheel odometry, and LiDAR, it is not limited to these specific applications. The core ideas and procedures of the pipeline can be potentially extended to more complex three-dimensional environments and other sensor modalities.

For example, cameras and 3D LiDARs estimate inter-frame relative poses by tracking visual or geometric features across viewpoints \cite{Woosik2024,Rehder16,LIU 2022}, microphone arrays can estimate the DOA in 3D space based on their spatial configurations \cite{Wang 2021,Jiang24}, and IMUs derive motion estimates via inertial pre-integration \cite{Woosik2024,Rehder16,LIU 2022}. These measurements can be integrated into the Fisher information matrix-based trajectory optimization framework (see (15)-(18)) to optimize the control points position in 3D space to improve data collection for parameter estimation. Therefore, the proposed method for active extrinsic calibration is potentially extendable to more complex 3D environments and a wider range of sensor modalities.

\section{Numerical Simulations}
A series of numerical experiments were conducted to evaluate the performance of the proposed observability-aware active calibration method for multi-sensor extrinsic parameters. First, the impact of varying levels of sensor measurement noise on the method performance was analyzed. Second, a comparison was made between the calibration results obtained using commonly employed fixed trajectories and those achieved with the proposed active calibration approach.

In the simulation experiments, we utilized a mobile robot equipped with a microphone array, LiDAR, and a wheel odometer. The robot frame coincided with the frame of the wheel odometer. The ground truth of the external parameters for the microphone array and LiDAR, relative to the wheel odometer, were set to $\left[0.3 \, \text{m}, 0.1 \, \text{m}, 60^\circ \right]$ and $\left[0.4 \, \text{m}, 0.2 \, \text{m}, 30^\circ \right]$, respectively. The robot motion was constrained within a $2.5 \text{m} \times 2.5 \, \text{m}$ area, and the stationary sound source used for calibrating the microphone array was placed at 
$\left[0 \, \text{m}, 3.6 \, \text{m} \right]$ in the global frame.

\begin{table}
    \caption{\label{tab:1}THE RMSE OF CALIBRATION RESULTS UNDER VARYING MEASUREMENT NOISE LEVELS: ANALYSIS OF 200 MONTE CARLO SIMULATIONS (BOLD MEANS BETTER)}
    \begin{center}
        \scalebox{1.0}{\begin{tabular}{ccccccc}  
            \toprule [1pt]
            \multicolumn{1}{l}{\multirow{3}{0.5cm}{Noise Levels}} &\multicolumn{3}{c}{Microphone Array} & \multicolumn{3}{c}{LiDAR} \\ 
            \cmidrule(lr){2-4}
            \cmidrule(lr){5-7}
            &  \makecell[c]{$^{W}\mathbf{t}_{Mx}$ \\ (m)} & \makecell[c]{$^{W}\mathbf{t}_{My}$ \\ (m)} & \makecell[c]{$^{W}\theta_M$\\ (deg)}& \makecell[c]{$^{W}\mathbf{t}_{Lx}$ \\ (m)} & \makecell[c]{$^{W}\mathbf{t}_{Ly}$ \\ (m)} & \makecell[c]{$^{W}\theta_L$\\ (deg)}\\  
            \midrule [1pt]
             Lv1  & \textbf{0.063} &	\textbf{0.044} &	\textbf{0.818} 	&\textbf{0.011} & \textbf{0.013} 	& \textbf{1.109}\\
            \rule{0pt}{10pt}
            Lv2 & 0.102 &	0.084 &	1.520 	&0.021 &	0.025 	&2.272  \\
            \rule{0pt}{10pt}
            Lv3 & 0.164 &	0.132 &	2.661 	&0.067 &	0.071 	&5.766        \\
            \rule{0pt}{10pt}
            Lv4 & 0.221 &	0.182 &	3.466 	&0.131 &	0.149 	&12.116       \\
            \rule{0pt}{10pt}
            Lv5 & 0.271 &	0.215 &	4.294   &0.189 &	0.234 	&21.487       \\
            \bottomrule [1pt]
        \end{tabular}}
    \end{center}
\end{table}

\subsection{Calibration under Varying Levels of Measurement Noise}
To investigate the impact of varying levels of sensor measurement noise on the performance of the proposed method, we employed the following strategy: first, Gaussian noises with varying levels were added to the sensor theoretical measurements; then, these noise-corrupted measurements were treated as actual data and used in the calibration process.

We specifically categorized the measurement noise of each sensor into five levels (Lv1 to Lv5) for analysis. For the wheel odometer, the noise standard deviation (std) was set to $2\%, 4\%, 6\%, 8\%,$ and $10\%$ of the distance traveled, respectively. For the microphone array, the measurement std was 2°, 4°, 6°, 8°, and 10° for estimating the DOA of the sound source, respectively. For the LiDAR, the noise std for relative displacement was set to $diag_2(0.005 \text{m}), diag_2(0.01 \text{m}), diag_2(0.03 \text{m}), diag_2(0.06 \text{m}),$ and $diag_2(0.1 \text{m})$, while for relative rotation, it was set to 0.5°, 2°, 3.5°, 5°, and 6.5°, respectively. These noise levels were designed to simulate typical sensor performance under varying environmental conditions.

For each set of noise levels, 200 Monte Carlo simulations of the active calibration were conducted, and the root mean square error (RMSE) was used as the evaluation metric to quantify the accuracy of the external parameter estimation. In all simulations, the initial sensor parameters were set to zero, and the initial a posteriori covariance matrix was initialized as an identity matrix.

The final calibration results are shown in Table \ref{tab:1}. It is evident that the estimation errors for the unknown parameters increase progressively with higher levels of measurement noise. Fig. \ref{fig:3} illustrates the trend of estimation errors with the number of observations for a representative dataset, along with the corresponding 3$\sigma$ confidence bounds. It shows that the 3$\sigma$ confidence bounds for all parameters decrease rapidly, indicating that the uncertainties associated with the unknown parameters in the sensor model diminish over time. This demonstrates that the proposed active calibration method effectively generates sufficient excitation to significantly enhance the observability of the unknown parameters. These results validate the effectiveness of the proposed method.

Fig. \ref{fig:4} illustrates the error distribution from Monte Carlo experiments. In the box plot, blue circles represent outliers identified using the 1.5 times interquartile range (IQR) criterion. The upper and lower black horizontal lines indicate the maximum and minimum values of the non-outlier errors, while the upper and lower edges of each box correspond to the 75th and 25th percentiles of the error distribution, respectively. The blue line in the center of the box indicates the median, reflecting the central tendency of the results. The orange triangles in the plot represent the corresponding RMSE. As the noise level increases, the error distribution of the calibration results widens, as indicated by a shift in the median, an expansion of the distribution, and an increase in the RMSE. This suggests that smaller measurement errors lead to more concentrated and stable calibration results.

Finally, we remark that calibration results under different levels of measurement noise can, to some extent, reflect the system performance across environments of varying scales within the sensors sensing range. Previous studies have shown that sensor perception performance often degrades as the environment size increases \cite{Jiang24,Cadena2016,wheel}, typically resulting in higher measurement noises. This observation indirectly suggests that our proposed method may achieve more accurate calibration in smaller environments.

\begin{table}
    \caption{\label{tab:2}
    THE RMSE OF CALIBRATION RESULTS UNDER DIFFERENT TRAJECTORY STRATEGIES: ANALYSIS OF 200 MONTE CARLO SIMULATIONS (BOLD MEANS BETTER)}
    \begin{center}
        \scalebox{1.0}{\begin{tabular}{ccccccc}  
            \toprule [1pt]
            \multicolumn{1}{l}{\multirow{3}{0.5cm}{Tajectory Strategies}} &\multicolumn{3}{c}{Microphone Array} & \multicolumn{3}{c}{LiDAR} \\ 
            \cmidrule(lr){2-4}
            \cmidrule(lr){5-7}
            &  \makecell[c]{$^{W}\mathbf{t}_{Mx}$ \\ (m)} & \makecell[c]{$^{W}\mathbf{t}_{My}$ \\ (m)} & \makecell[c]{$^{W}\theta_M$\\ (deg)}& \makecell[c]{$^{W}\mathbf{t}_{Lx}$ \\ (m)} & \makecell[c]{$^{W}\mathbf{t}_{Ly}$ \\ (m)} & \makecell[c]{$^{W}\theta_L$\\ (deg)}\\ 
            \midrule [1pt]
            active(Ours)  & 0.063 &	\textbf{0.044} &	\textbf{0.818} 	&\textbf{0.011} & \textbf{0.013} 	& \textbf{1.109}\\
            \rule{0pt}{10pt}
            figure-8 & \textbf{0.060} &	\textbf{0.044} & 4.772 & 0.030 &	0.053 &	4.780          \\
            \rule{0pt}{10pt}
            circle & 0.073 &	0.069 &	1.692 & 0.565 &	0.264 &	22.001         \\
            \rule{0pt}{10pt}
            line & 0.119   &	0.126 &	2.843 & 0.400 &	0.200 &	2.763           \\
            \bottomrule [1pt]
        \end{tabular}}
    \end{center}
\end{table}

\begin{table}
    \caption{\label{tab:init}THE RMSE OF CALIBRATION RESULTS UNDER VARYING INITIALIZATION ERRORS: ANALYSIS OF 200 MONTE CARLO SIMULATIONS (BOLD MEANS BETTER)}
    \begin{center}
        \scalebox{1.0}{\begin{tabular}{ccccccc}  
            \toprule [1pt]
            \multicolumn{1}{l}{\multirow{3}{0.5cm}{Initial Errors}} &\multicolumn{3}{c}{Microphone Array} & \multicolumn{3}{c}{LiDAR} \\ 
            \cmidrule(lr){2-4}
            \cmidrule(lr){5-7}
            &  \makecell[c]{$^{W}\mathbf{t}_{Mx}$ \\ (m)} & \makecell[c]{$^{W}\mathbf{t}_{My}$ \\ (m)} & \makecell[c]{$^{W}\theta_M$\\ (deg)}& \makecell[c]{$^{W}\mathbf{t}_{Lx}$ \\ (m)} & \makecell[c]{$^{W}\mathbf{t}_{Ly}$ \\ (m)} & \makecell[c]{$^{W}\theta_L$\\ (deg)}\\  
            \midrule [1pt]
             Err1  & \textbf{0.097} &	\textbf{ 0.092} & \textbf{1.678}&	\textbf{0.018} 	&\textbf{0.025} & \textbf{2.585} \\
            \rule{0pt}{10pt}
            Err2 & 0.109 &	0.100 &	2.041 	&0.022 &	0.030 	&3.395  \\
            \rule{0pt}{10pt}
            Err3 & 0.279 &	0.344 &	5.344 	&0.020 &	0.041 	& 5.056        \\
            \rule{0pt}{10pt}
            Err4 & 0.459 &	0.505 &	7.869 	&0.091 &	0.089 	& 17.359       \\
            \bottomrule [1pt]
        \end{tabular}}
    \end{center}
\end{table}

\subsection{Calibration under Different Trajectory Strategies}
Next, we compared the accuracy of the proposed method with three different fixed trajectories, i.e., a line trajectory, a circular trajectory, and a figure-eight trajectory from the origin to (2m, 2m). In the experiments, the measurement noise level of sensors is set to LV1 (refer to Section V.A). 
For each configuration, we performed 200 Monte Carlo simulations. As in the setup described in Section V.A, the initial guesses of the unknown parameters were all set to zero, the initial a posteriori covariance matrix was initialized as an identity matrix, and the RMSE served as the evaluation metric.

The final calibration results, presented in Table \ref{tab:2}, demonstrate that the proposed calibration method generally achieves significantly higher accuracy in parameter estimation. While the figure-eight trajectory shows a slight advantage in calibrating the position of the microphone array, it performs poorly in estimating sensor orientations. Similarly, the linear and circular trajectories exhibit poor performance, particularly in estimating the external parameters of the LiDAR.

As shown in Fig. \ref{fig:5}, the proposed method significantly outperforms the other fixed trajectories in terms of the convergence speed of the 3$\sigma$ confidence bounds, effectively reducing the uncertainty in parameter estimation. 
Fig. \ref{fig:6} further highlights that the proposed method achieves a smaller error distribution, significantly lower RMSE and fewer outliers. 
Note that the linear trajectory fails to provide sufficient excitation for the measurement model due to the parallel motion between the wheel odometer and the LiDAR. This limitation hinders the convergence of the LiDAR position parameters, as evidenced by the nearly constant 3$\sigma$ boundaries and the minimal changes in the corresponding box plots.

\subsection{Calibration under Different Initialization Errors}
To evaluate the impact of initial guess errors on calibration results, we conducted experiments by adding Gaussian noise of varying magnitudes to the ground truth. Specifically, under the LV2 measurement noise level (see Section V-A), we considered four levels of initial guess errors, denoted as Err1, Err2, Err3, and Err4, corresponding to standard deviations of (0.2 m, 0.2 m, 10°), (0.5 m, 0.5 m, 30°), (0.8 m, 0.8 m, 50°), and (1.0 m, 1.0 m, 70°) for the sensor position and orientation, respectively. For each error level, we randomly generated initial guesses and performed 200 Monte Carlo simulations, using the RMSE to assess the calibration accuracy.

Table \ref{tab:init} presents the experimental results. As the initial guess errors increased, the final calibration errors also tended to grow. Furthermore, we classified a trial as an outlier if any sensor position error exceeded 3 meters or any orientation error exceeded 90 degrees. Across the four error levels, the number of outliers observed in the Monte Carlo simulations was 0, 5, 5, and 17, respectively. These results collectively demonstrate that providing a good initial guess is crucial for ensuring the accuracy and robustness of the proposed algorithm.

\begin{figure}[t]
	\centering
	\begin{minipage}{0.43\linewidth}
		\centering
		\subfigure[]{\includegraphics[width=1\linewidth]{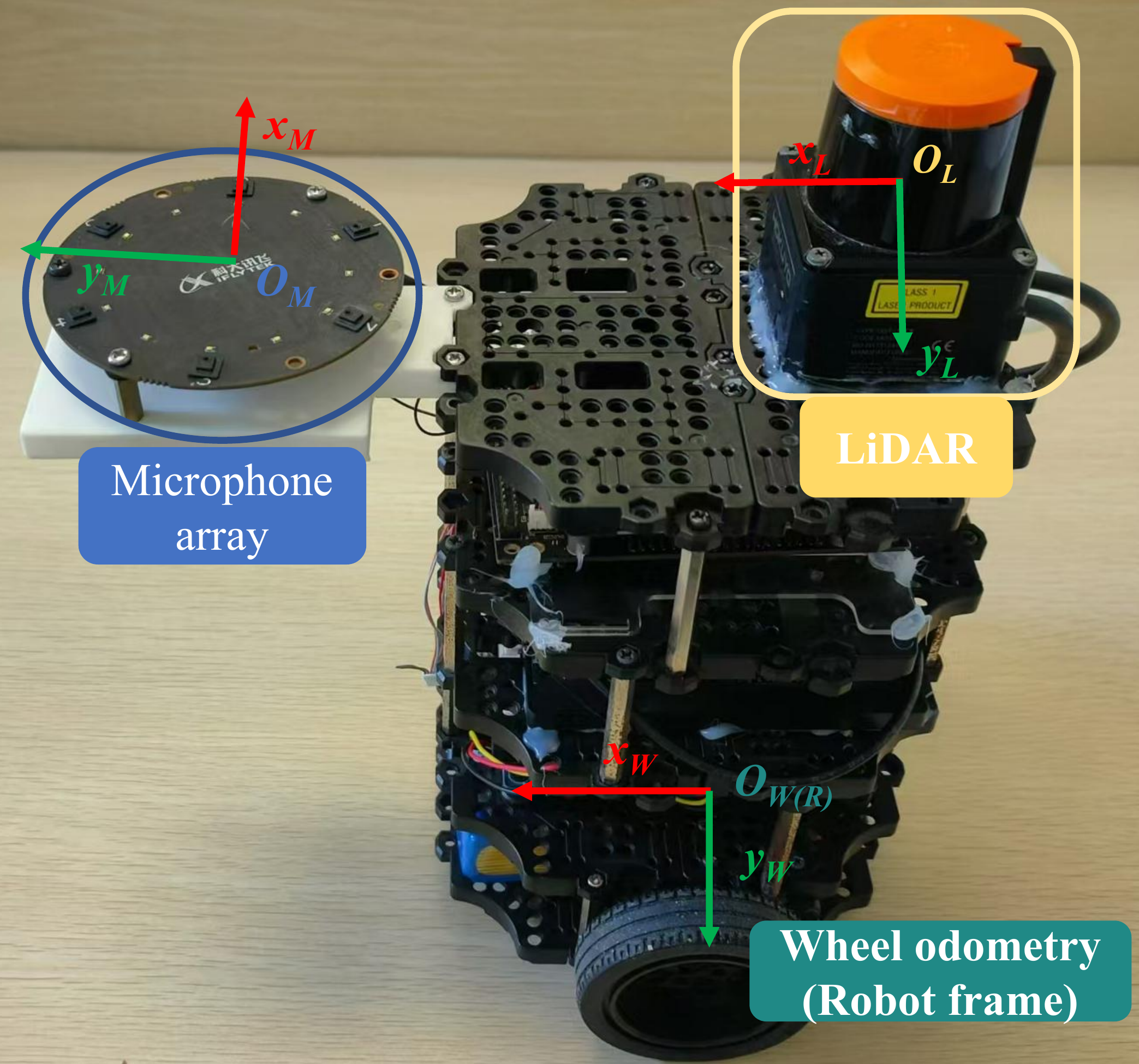}}
	\end{minipage}
	\begin{minipage}{0.53\linewidth}
		\centering
		\subfigure[]{\includegraphics[width=1\linewidth]{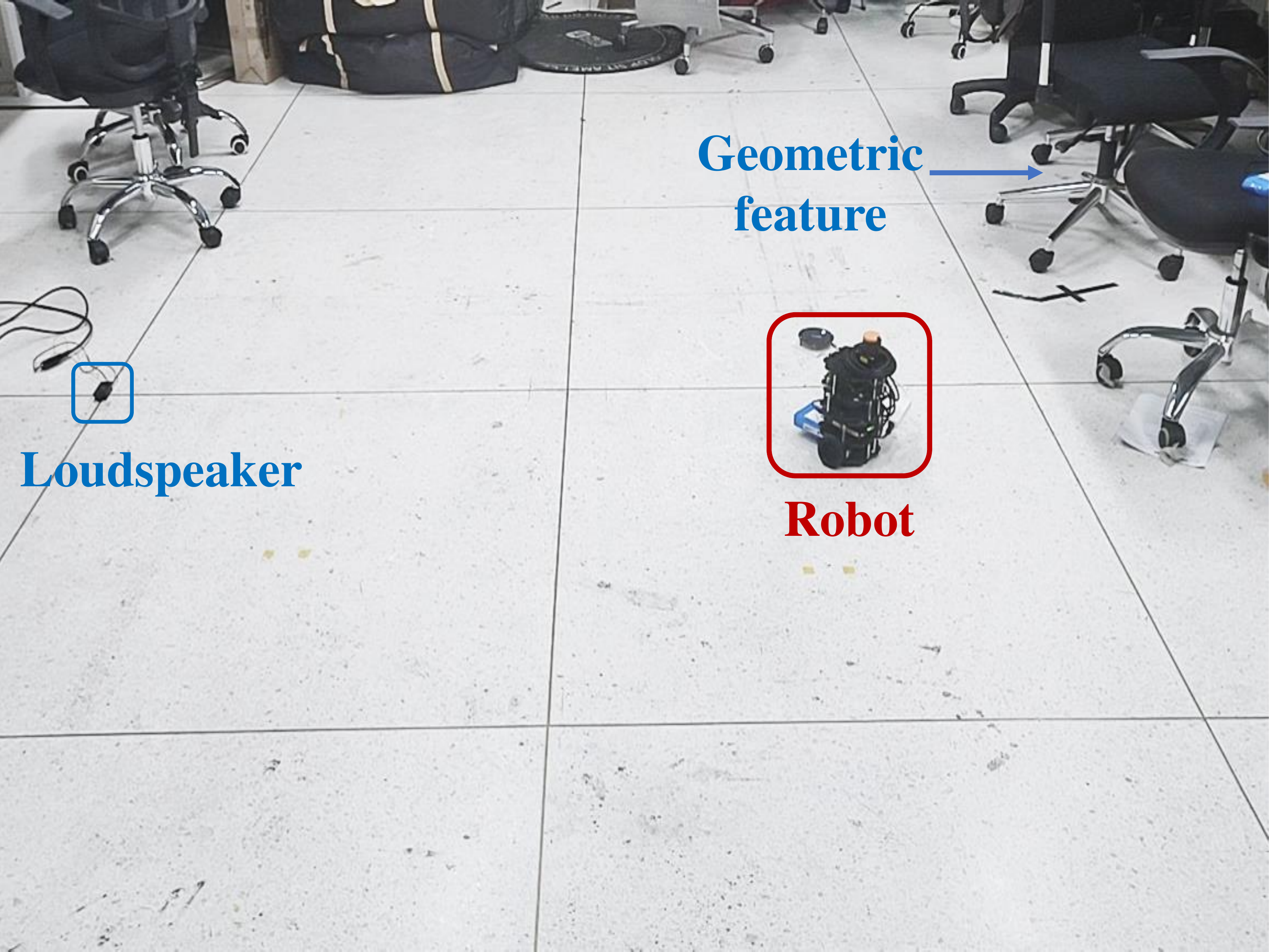}}
	\end{minipage}
    	\caption{Real-world multi-sensor calibration environment setup. (a)  Turtlebot3 robot with Multi-sensors. (b) Typical physical scenario. }
    \label{fig:exp_scene}
\end{figure}

\begin{figure*}[htbp]
	\centering
	\begin{minipage}{0.32\linewidth}
		\centering
		\subfigure[]{\includegraphics[width=0.8\linewidth]{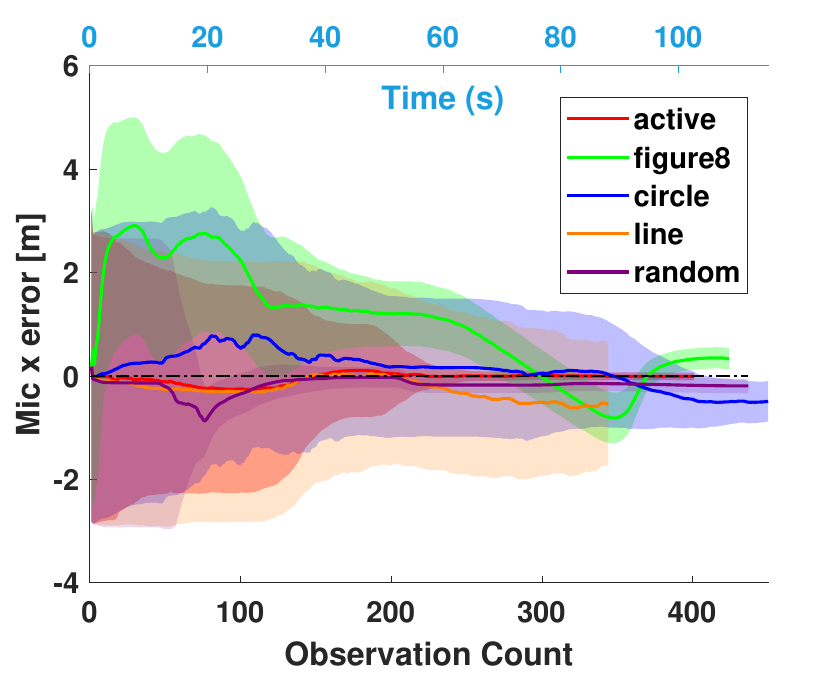}}
	\end{minipage}
	\begin{minipage}{0.32\linewidth}
		\centering
		\subfigure[]{\includegraphics[width=0.8\linewidth]{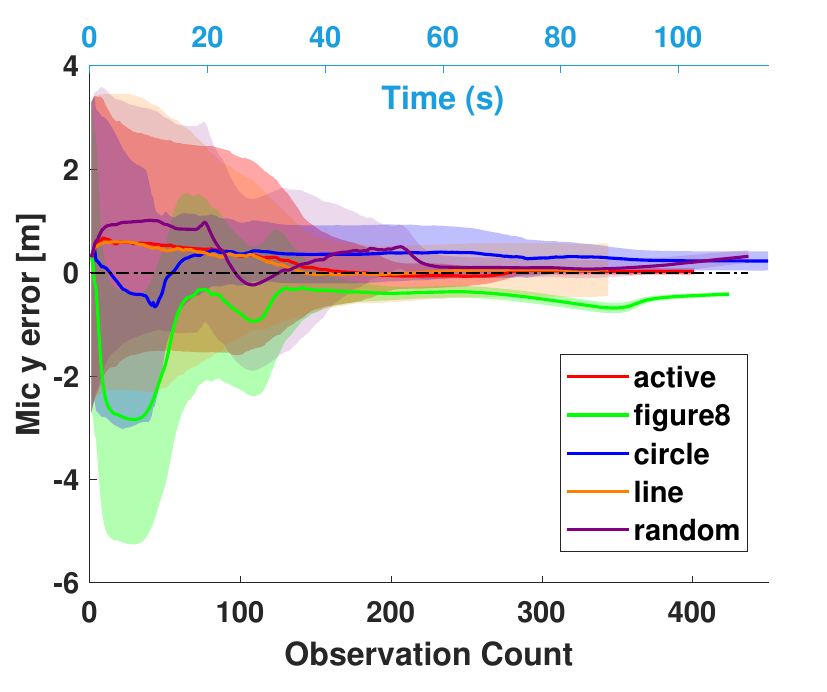}}
	\end{minipage}
	\begin{minipage}{0.32\linewidth}
		\centering
		\subfigure[]{\includegraphics[width=0.8\linewidth]{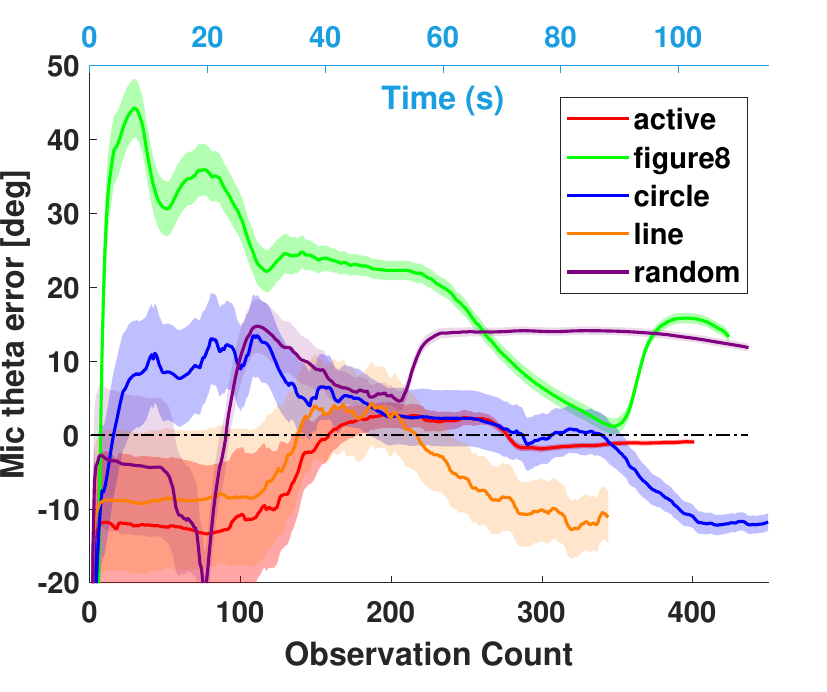}}
	\end{minipage}
    \qquad
	\begin{minipage}{0.32\linewidth}
		\centering
		\subfigure[]{\includegraphics[width=0.8\linewidth]{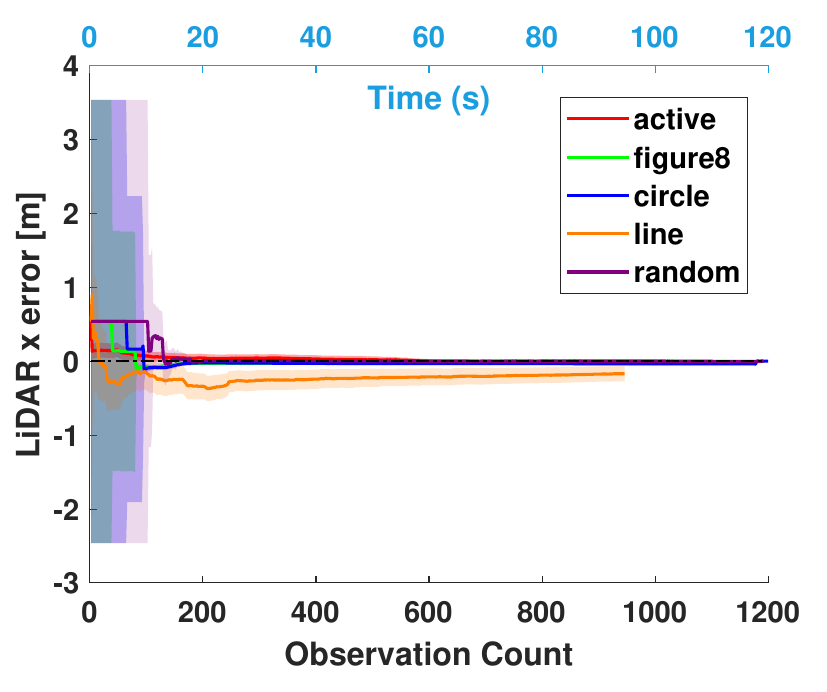}}
	\end{minipage}
	\begin{minipage}{0.32\linewidth}
		\centering
		\subfigure[]{\includegraphics[width=0.8\linewidth]{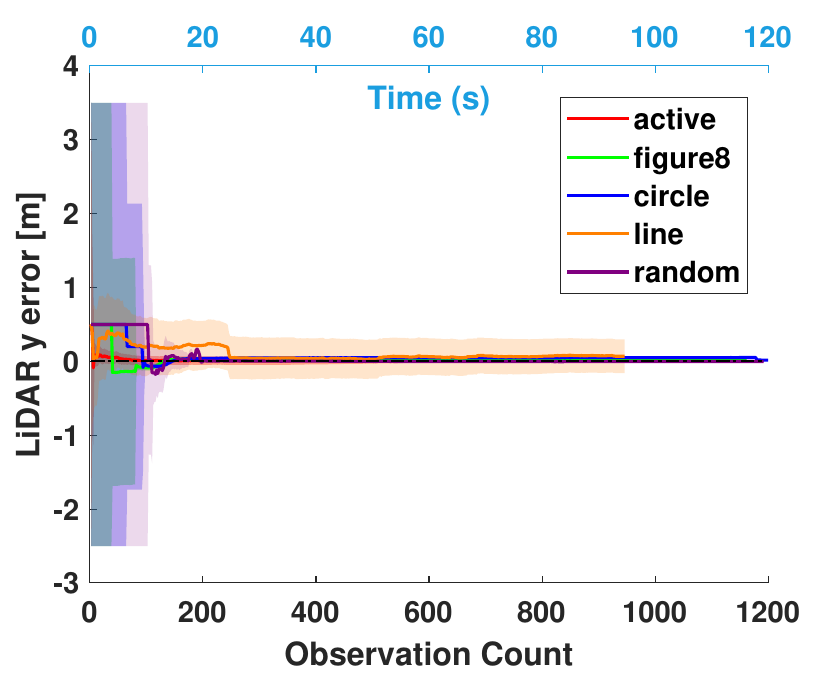}}
	\end{minipage}
    \begin{minipage}{0.32\linewidth}
		\centering
		\subfigure[]{\includegraphics[width=0.8\linewidth]{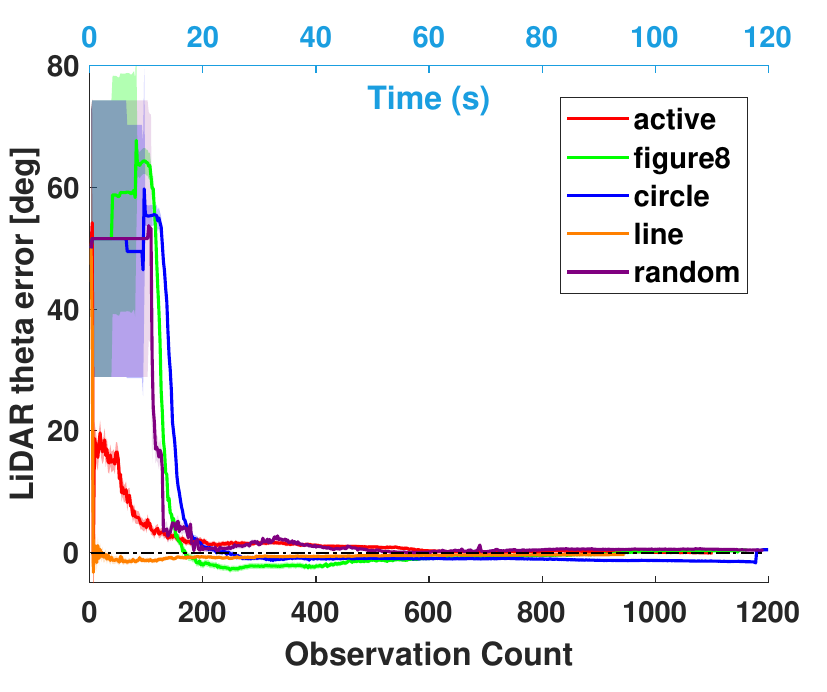}}
	\end{minipage}
	\caption{The solid lines represent the estimation errors of the calibration results in real-world (obtained from commonly employed fixed trajectories and the proposed active calibration method), while the shaded regions denote the corresponding 3$\sigma$ bounds.} 
    \label{fig:8}
\end{figure*}
\section{Real-World Experiments}
To further evaluate and validate the proposed framework, we conducted experiments using a real mobile robot platform. The experiments were performed on a Turtlebot3 robot equipped with a Jetson Xavier NX embedded computer and various sensors rigidly mounted, as illustrated in Fig. \ref{fig:exp_scene}(a). These sensors include an iFLYTEK M160C microphone array consisting of six microphones uniformly arranged in a circle, each with a 32-bit sampling depth and a 16 kHz sampling rate; a HOKUYO UST-10LX LiDAR with a sampling rate of 43.2 kHz and an angular resolution of 0.25°; and a wheel odometer integrated with two 12-bit encoders attached to two Dynamixel XL430-W250 motors, respectively. 

\subsection{Experimental Setup}
Prior to the experiments, we used a protractor and a vernier caliper to measure the relative positions and orientations of the microphone array and the LiDAR w.r.t. the geometric center of the wheel odometry frame. The measured values were $\left[0.12 \, \text{m}, 0 \, \text{m}, 270^\circ \right]$ for the microphone array and $\left[-0.04 \, \text{m}, 0 \, \text{m}, -1.6^\circ \right]$ for the LiDAR. These measurements served as ground truth for evaluating the calibration accuracy.

The experiments were conducted in a room measuring 9m×6m, with the robot trajectory planning  restricted to a 2.5m×2.5m area. Since the robot lacks strong perception capabilities prior to calibration, limiting the motion space is both practical and safe \cite{Preiss17}. The starting position of the robot platform was designated as the origin, and a stationary loudspeaker was placed on the ground. The loudspeaker emitted a continuous chirp signal ranging from 3000 Hz to 5000 Hz with a period of 128 ms throughout the experiment. We carried out active calibration experiments by utilizing signals emitted by the loudspeaker, geometric features from the environment, and its own motion data, as illustrated in Fig. \ref{fig:exp_scene}(b). 

In our experiments, the robot inevitably encounters self-motion noise from servo and the LiDAR motor as well as environmental background noise generated. Given that the theoretical DOA measurements from the microphone array are expected to vary smoothly as the robot follows the B-spline trajectory, we apply an extended Kalman filter to suppress outliers in the DOA estimates caused by such disturbances. This filtering strategy improves both the stability and accuracy of DOA estimation. 


\subsection{Results}
The system adopts a parallel architecture to enable real-time sensor data acquisition, processing, and motion planning. Each 256 ms audio segment is used for DOA estimation, while 100 ms of LiDAR data is used to estimate relative pose changes, with average processing times of 35 ms and 2 ms, respectively. Each sensor is then applied an EKF separately, which updates the sensor parameters within just 1 ms. Since the processing time for each modality is shorter than its corresponding acquisition period, all sensor data can be handled in real time without delay. Trajectory optimization, however, is more computationally intensive, requiring an average of 3.5 s per B-spline segment. To maintain seamless system operation, the next trajectory segment is optimized once the robot start the current segment. These treatments collectively ensure that the robot can carry out active calibration tasks online.

To evaluate the effectiveness of the proposed error compensation strategy in real-world settings, we conducted the following experiments: 1. active (Ours), i.e., using LiDAR-based pose estimation to replace wheel odometry (once the LiDAR parameters had converged); 2. active (Wheel), using wheel odometry for pose estimation of the robot. For each configuration, we repeated the experiment three times and used RMSE as the evaluation metric. As shown in Table \ref{tab:3}, the result demonstrates that the proposed compensation significantly reduces the estimation error of microphone parameters compared to the uncompensated case.

Subsequently, similar to Section V.B., we compared the accuracy of the proposed active calibration method with different fixed trajectories, including random, linear, circular, and figure-eight patterns, which are commonly used in calibration scenarios. The random trajectory was generated by fitting a B-spline curve through randomly sampled control points. Each configuration was conducted with three experimental repetitions. The final calibration results, shown in Table \ref{tab:3}, demonstrate that the proposed method achieves better performance, particularly in estimating sensor orientation and microphone array position. 
Take one typical calibration dataset as an example, the proposed active calibration method significantly reduced the 3$\sigma$ confidence bounds of the parameters, converged faster, and achieved more accurate results with fewer observations and shorter time compared to the fixed-trajectory calibration, as shown in Fig. \ref{fig:8}.


\begin{table}[t]
    \caption{\label{tab:3}THE RMSE OF CALIBRATION RESULTS OF DIFFERENT CONFIGURATION IN REAL-WORLD (BOLD MEANS BETTER)}
    \begin{center}
        \scalebox{1.0}{\begin{tabular}{ccccccc}  
            \toprule [1pt]
            \multicolumn{1}{l}{\multirow{3}{0.5cm}{Tajectory Strategies}} &\multicolumn{3}{c}{Microphone Array} & \multicolumn{3}{c}{LiDAR} \\ 
            \cmidrule(lr){2-4}
            \cmidrule(lr){5-7}
            &  \makecell[c]{$^{W}\mathbf{t}_{Mx}$ \\ (m)} & \makecell[c]{$^{W}\mathbf{t}_{My}$ \\ (m)} & \makecell[c]{$^{W}\theta_M$\\ (deg)}& \makecell[c]{$^{W}\mathbf{t}_{Lx}$ \\ (m)} & \makecell[c]{$^{W}\mathbf{t}_{Ly}$ \\ (m)} & \makecell[c]{$^{W}\theta_L$\\ (deg)}\\ 
            \midrule [1pt]
            active (Ours)  & \textbf{0.006} &	\textbf{0.009} &	\textbf{0.611} & 0.003&	0.015 	& \textbf{0.118} \\
            \rule{0pt}{10pt}
            {active (Wheel)}  & {0.072} &	{0.030} &	{3.173}  & {\textbf{0.002}} & {0.015} & {0.163}    \\
            \rule{0pt}{10pt}
            figure-8 & 0.203& 	0.243 &	8.043 & 0.007 &	0.015 	& 0.361    \\
            \rule{0pt}{10pt}
            circle & 0.582  &	0.312 &	12.852& 0.013 &	0.018 	& 0.706     \\
            \rule{0pt}{10pt}
            line & 0.685 	&   0.150 &	13.922&  0.133 &	0.350& 	1.188            \\
            \rule{0pt}{10pt}
            {random} & 0.272 	&   {0.285} &	9.932&  {0.006} &	\textbf{0.010} & 	{0.304}            \\
            \bottomrule [1pt]
        \end{tabular}}
    \end{center}
\end{table}

\section{Limitations and Discussions}
Simulation and real-world results demonstrate that the proposed method exhibits strong robustness and significantly improves calibration accuracy and convergence speed compared to widely used calibration trajectories. It should be noted that the calibration performance heavily depends on the accuracy of sensor measurements.

For the microphone array, our simulations show that the calibration framework relies heavily on accurate DOA estimation, while real-world experiments indicate that the method tolerates certain noise, such as self-motion noise and environmental background noise. However, in more complex acoustic environments, such as those with strong reverberations, multiple interfering sources, or intermittent signals, there are greater challenges and may degrade the performance. In such cases, integrating more advanced signal processing and DOA estimation techniques, such as those discussed in \cite{Cao24, AnTRO,Wang 2021}, would be crucial essential to maintain calibration accuracy.

For LiDAR, the proposed method depends on the availability of rich scene features to obtain accurate relative pose measurements. In environments with sparse structures or limited field of view, insufficient point cloud information may prevent reliable pose estimation \cite{LIU 2022}, thereby reducing the method effectiveness (see the discussed in Section V-A). To address this issue, incorporating state-of-the-art point cloud registration techniques \cite{fastlio,Pomerleau F} becomes essential.

Moreover, thanks to the flexibility of our trajectory representation and optimization framework, arbitrary trajectory segments can be adaptively adjusted during calibration while respecting the robot’s motion constraints. This capability offers great potential for extending the method to more complex scenarios, such as environments with obstacles. For applications requiring higher real-time performance, combining local and global trajectory planning strategies could further enhance active calibration capabilities under challenging conditions.



\section{Conclusion}
In this study, we have presented an observability-aware active calibration method for extrinsic calibration of ground robotic systems equipped with exteroceptive sensors (microphone arrays and LiDAR) and proprioceptive sensors (wheel odometry). 
The proposed framework utilizes the FIM to evaluate the observability of system parameters and integrate this metric into a trajectory generation framework based on B-spline curves. The robot autonomously plans and dynamically adjusts its trajectory to maximize the minimal eigenvalue of the FIM, thereby improving the observability of unknown parameters. During the data collection process, an EKF updates the extrinsic parameters of all sensors in real-time. Extensive simulation and real-world experiments demonstrate that the proposed method significantly enhances calibration accuracy and accelerates convergence compared to conventional data collection strategies that adopt fixed trajectories, including circular, and figure-eight patterns. In our future work, we aim to extend the proposed method to more diverse robotic platforms (e.g., drones and legged robots), incorporating a broader range of sensor modalities (e.g., camera and IMU) and operating in more complex environments.




\begin{appendices}
\section*{Appendix\\Jacobian of Sensor Models}\label{secA1}
Given the observation model (see (\ref{expression_DOA-1}) and (\ref{expression_lidar})), we can derive their Jacobian matrices w.r.t. their respective states $\mathbf{\psi}_M$ and $\mathbf{\psi}_L$. For the microphone array, we denote its position in global frame at time instance $k$, $k = 1, . . . , K,$ as:
\begin{equation}
^{G}\mathbf{t}_{M,k} =\left[^{G}t_{M,k}^x,^{G}t_{M,k}^y\right]= ^{G}\mathbf{t}_k+^{G}\mathbf{R}_{k} \cdot ^{W}\mathbf{t}_{M}.
\end{equation}
Denote the sound source position in the global frame as:
\begin{equation}
\mathbf{s}=\left[s^{x};s^{y}\right].\label{eq:Source location}
\end{equation}
The distance between the sound source $\mathbf{s}$ and the microphone array at time instance $k$ can be computed as:
\begin{equation}
d_{k}=\sqrt{(\Delta x_{M\_k})^{2}+(\Delta y_{M\_k})^{2}},
\end{equation}
where
\begin{equation}
\Delta x_{M\_k}=s^{x}-^{G}t_{M,k}^x,\text{ }
\Delta y_{M\_i}=s^{y}-^{G}t_{M,k}^y.
\end{equation}
Then the Jacobian matrix of the microphone array measurement model $\mathbf{d}_{k}$ in (\ref{expression_DOA-1}) w.r.t. $\mathbf{\psi}_M$ could be expressed as:
\begin{equation}
\frac{\partial\mathbf{d}_{k}}{\partial\psi_{M}}=\left[\frac{\partial\mathbf{d}_{k}}{\partial^{W}\mathbf{t}_{M}},\frac{\partial\mathbf{d}_{k}}{\partial^{W}\theta_{M}}\right],
\end{equation}
where
\begin{equation}
\frac{\partial\mathbf{d}_{k}}{\partial^{W}\mathbf{t}_{M}}=\frac{\left(^{G}\mathbf{R}_{k}\cdot^{W}\mathbf{R}_{M}\right)^{\top}}{d_{k}^{3}}\left(-^{G}\mathbf{R}_{k}d_{k}^{2}+
\mathbf{A}{}^{G}\mathbf{R}_{k}\right)
\end{equation}
with 
\begin{equation}
\mathbf{A} = 
\begin{bmatrix}
\begin{smallmatrix}
(\Delta x_{M\_k})^{2} & \Delta x_{M\_k}\Delta y_{M\_k}\\
\Delta x_{M\_k}\Delta y_{M\_k} & (\Delta y_{M\_k})^{2}
\end{smallmatrix}
\end{bmatrix},
\end{equation}
and
\begin{equation}
\frac{\partial\mathbf{d}_{k}}{\partial^{W}\theta_{M}}=\frac{\partial{}^{W}\mathbf{R}_{M}^{\top}}{\partial^{W}\theta_{M}}{}^{G}\mathbf{R}_{k}^{\top}\frac{\mathbf{s}-^{G}\mathbf{t}_{M,k}}{d_{k}}
\end{equation}
with
\begin{equation}
\frac{\partial{}^{W}\mathbf{R}_{M}^{\top}}{\partial^{W}\theta_{M}}=
\left[\begin{array}{cc}
-\sin\left({}^{W}\theta_{M}\right)  & \cos\left({}^{W}\theta_{M}\right)  \\
-\cos\left({}^{W}\theta_{M}\right)   & -\sin\left({}^{W}\theta_{M}\right)
\end{array}\right].
\end{equation}

The Jacobian matrix of the LiDAR measurement model $\Delta\mathbf{t}_{L,k}$ and $\Delta \theta_{L,k}$ in (\ref{expression_lidar}) w.r.t. $\mathbf{\psi}_L$ could be expressed as:
\begin{equation}
\left[\begin{array}{c}
\frac{\partial\Delta\mathbf{t}_{L,k}}{\partial\psi_{L}} \\
\frac{\partial\Delta \theta_{L,k}}{\partial\psi_{L}}
\end{array}\right]=
\left[\begin{array}{cc}
\frac{\partial\Delta\mathbf{t}_{L,k}}{\partial^{W}\mathbf{t}_{L}} & \frac{\partial\Delta\mathbf{t}_{L,k}}{\partial^{W}\theta_{L}}\\
\mathbf{0} & 0
\end{array}\right],
\end{equation}
where
\begin{equation}
\frac{\partial\Delta\mathbf{t}_{L,k}}{\partial^{W}\mathbf{t}_{L}}=\left(^{G}\mathbf{R}_{k-1}\cdot^{W}\mathbf{R}_{\mathit{L}}\right)^{\top}\left(^{G}\mathbf{R}_{k}-^{G}\mathbf{R}_{k-1}\right),
\end{equation}
and
\begin{equation}
\frac{\partial\Delta\mathbf{t}_{L,k}}{\partial^{W}\mathbf{\theta}_{L}}=\frac{\partial{}^{W}\mathbf{R}_{\mathit{L}}^{\top}}{\partial^{W}\mathbf{\theta}_{L}}{}^{G}\mathbf{R}_{k-1}^{\top}\left(\Delta^{G}\mathbf{R}_{W,k}\cdot^{W}\mathbf{t}_{L}+\Delta\mathbf{t}_{k-1}^{k}\right),
\end{equation}
where 
\begin{equation}
\frac{\partial{}^{W}\mathbf{R}_{L}^{\top}}{\partial^{W}\theta_{L}}=
\left[\begin{array}{cc}
-\sin\left({}^{W}\theta_{L}\right)  & \cos\left({}^{W}\theta_{L}\right)  \\
-\cos\left({}^{W}\theta_{L}\right)   & -\sin\left({}^{W}\theta_{L}\right)
\end{array}\right].
\end{equation}
\end{appendices}

\begin{IEEEbiography}    [{\includegraphics[width=1in,height=1.25in,clip,keepaspectratio]
    {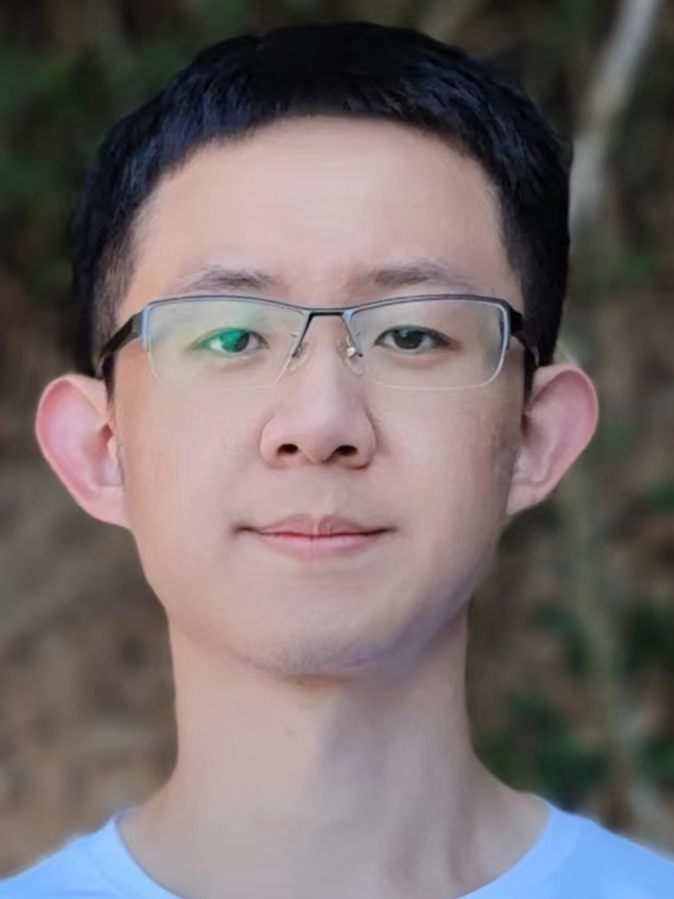}}]{Jiang Wang}
received the B.Eng. in Electrical Engineering and Automation from the Shenyang Agricultural University, Shenyang, China, in 2020, the M.Eng. in Electronic Science and Technology from the Southern University of Science and Technology, Shenzhen, China, in 2024. Since October 2024, he has been working towards the Ph.D. degree in Systems and Control Engineering with the Institute of Science Tokyo (formerly Tokyo Tech), Tokyo, Japan. His major research interests include sensor calibration, robot audition, SLAM, sensor fusion. 
\end{IEEEbiography}
\vspace{-2mm}
\begin{IEEEbiography}
[{\includegraphics[width=1in,height=1.25in,clip,keepaspectratio]
    {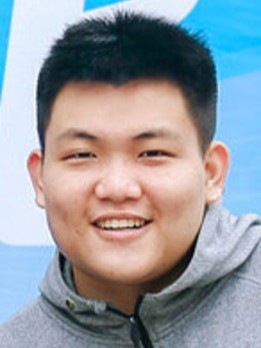}}]{Yaozhong Kang}
is currently working toward the B.Eng. degree in Automation with the Southern University of Science and Technology, Shenzhen, China. His research interests include mobile robot navigation, sensor calibration, and simultaneous localization and mapping. 
\end{IEEEbiography}

\begin{IEEEbiography}    [{\includegraphics[width=1in,height=1.25in,clip,keepaspectratio]
    {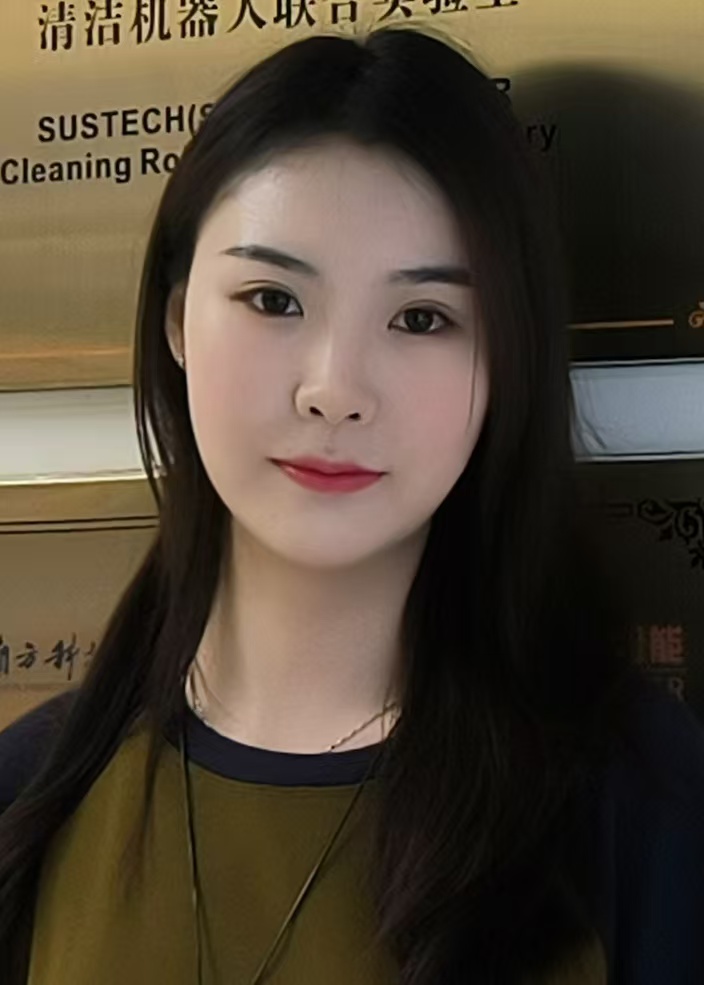}}]{Linya Fu} received the B.Eng. degree in Automation from China University of Mining and Technology, Xuzhou, China, in 2022. She is currently pursuing the M.Eng. degree in Intelligent Manufacturing and Robotics at the Southern University of Science and Technology, Shenzhen, China. Her research interests include robot audition, sound source localization, deep learning, and multi-modal perception. 
\end{IEEEbiography}

\begin{IEEEbiography}
[{\includegraphics[width=1in,height=1.25in,clip,keepaspectratio]{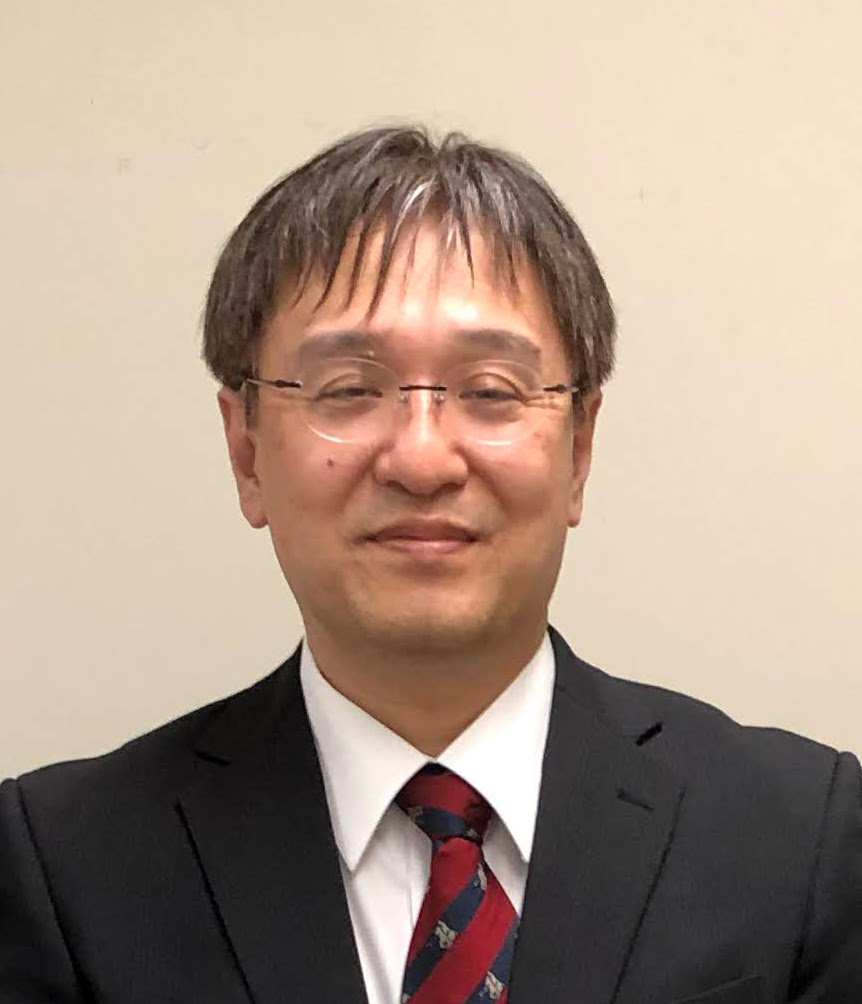}}]{Kazuhiro Nakadai} (Fellow, IEEE) \footnotesize{received his B.E. in Electrical Engineering in 1993, M.E. in Information Engineering in 1995, and Ph.D. in Electrical Engineering in 2003, all from the University of Tokyo. From 1995 to 1999, he worked as a systems engineer at Nippon Telegraph and Telephone Corporation. He then served as a researcher with the Kitano Symbiotic Systems Project, ERATO, JST from 1999 to 2003. From 2003 to 2022, he was a principal scientist at Honda Research Institute Japan, Co., Ltd. He is currently a professor in the Department of Systems and Control Engineering at the Institute of Science Tokyo (formerly Tokyo Institute of Technology), Tokyo, Japan. He has also held concurrent positions at Tokyo Institute of Technology: as a visiting associate professor (2006–2010), visiting professor (2011–2017), and specially appointed professor (2017–2022). Additionally, he was a guest professor at Waseda University from 2011 to 2018. His research interests include artificial intelligence, robotics, signal processing, computational auditory scene analysis, multimodal integration, and robot audition. He has served as an executive board member of the \textit{Japanese Society for Artificial Intelligence (JSAI)} during 2015–2017 and 2024–2026, and of the \textit{Robotics Society of Japan (RSJ)} during 2017–2018 and 2025–2026.}
\end{IEEEbiography}

\begin{IEEEbiography}
[{\includegraphics[width=1in,height=1.25in,clip,keepaspectratio]{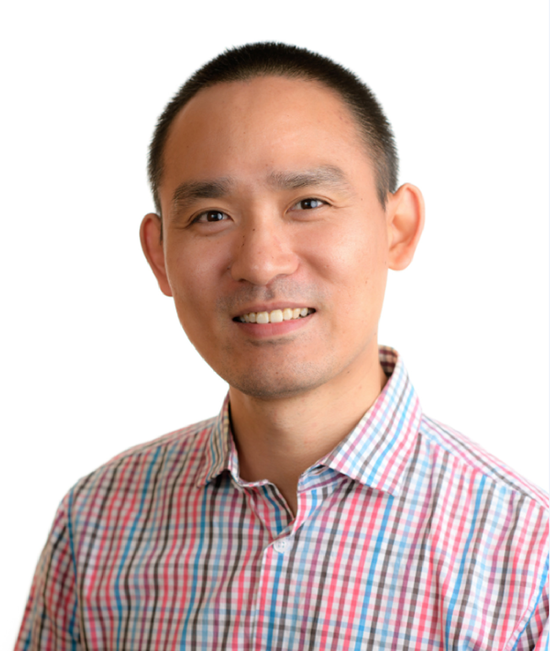}}]{He Kong} received 
the Ph.D. degree in Electrical Engineering from the University of Newcastle, Australia. He was a Research Fellow at the Australian Centre for Field Robotics, the University of Sydney, Australia, during 2016–2021. In early 2022, he joined the Southern University of Science and Technology, Shenzhen, China, where he is currently an Associate Professor. His research interests include robotic perception, robot audition, state estimation, and control applications. He is currently serving on the editorial board of \textit{IEEE Robotics and Automation Letters}, \textit{IEEE Robotics and Automation Magazine}, \textit{IEEE Sensors Letters}, \textit{International Journal of Adaptive Control and Signal Processing, Proceedings of the IMechE-Part I: Journal of Systems and Control Engineering, Journal of Climbing and Walking Robots}. He has also served as an Associate Editor on the IEEE CSS Conference Editorial Board and for the IEEE RAS flagship conferences such as the IEEE ICRA, IEEE/RSJ IROS, IEEE CASE, etc. As a co-recipient, he has received several awards, including the Best Paper Award at the 14-th International Conference on Indoor Positioning and Indoor Navigation in 2024, the Outstanding Poster Prize at the 5-th Annual Conference of China Robotics Society in 2024, a Finalist for the Young Author Award at the 1-st IFAC Workshop on Robot Control in 2019.
\end{IEEEbiography}
\end{document}